\documentclass[sigconf]{acmart}
\AtBeginDocument{%
  }

\copyrightyear{2027}
\acmYear{2027}
\setcopyright{cc}
\setcctype{by}
\acmConference[WSDM '27]{Proceedings of the Twentieth ACM International Conference on Web Search and Data Mining}{February 15--19, 2027}{Hong Kong, Hong Kong}
\acmBooktitle{Proceedings of the Twentieth ACM International Conference on Web Search and Data Mining (WSDM '27), February 15--19, 2027, Hong Kong, Hong Kong}
\acmDOI{10.1145/3862322.3863633}
\acmISBN{979-8-4007-3117-4/2027/02}

\acmSubmissionID{1878}

\usepackage{mathtools}      
\usepackage{subcaption}     
\usepackage{colortbl}       
\usepackage{makecell}
\usepackage{enumitem}
\usepackage{bm}
\usepackage{multirow}
\usepackage{diagbox}        
\usepackage{longtable}      
\usepackage{listings}
\usepackage{tcolorbox}      
\usepackage{cleveref}       
\crefformat{footnote}{#2\footnotemark[#1]#3}

\definecolor{dkgreen}{rgb}{0,0.6,0}
\definecolor{gray}{rgb}{0.5,0.5,0.5}
\definecolor{mauve}{rgb}{0.58,0,0.82}

\lstnewenvironment{pythontable}
  {\lstset{language=Python, frame=none,
  aboveskip=-1.5mm,
  belowskip=-2mm,
  columns=fullflexible,  
  basicstyle=\tiny\ttfamily,
    }%
  }
  {}

\makeatletter
\newcommand{\thickhline}{%
    \noalign {\ifnum 0=`}\fi \hrule height 1pt
    \futurelet \reserved@a \@xhline
}
\newcolumntype{"}{@{\hskip\tabcolsep\vrule width 1pt\hskip\tabcolsep}}
\makeatother

\makeatletter
\newcommand{\labelname}[1]{
  \def\@currentlabelname{#1}}%
\makeatother

\usepackage{microtype}
\DeclareMathOperator*{\argmax}{argmax}

\DeclareCaptionFormat{cont}{#1 (cont.)#2#3\par}

\begin{document}

\title{ReFrame: Automatic Problem Transformation for LLM-Based Heuristic Design}

\author{Nguyen Thach}
\affiliation{%
  \institution{University of Nebraska-Lincoln}
  \city{Lincoln, NE}
  \country{USA}}
\email{nate.thach@huskers.unl.edu}

\author{Nathan Huynh}
\affiliation{%
  \institution{University of Nebraska-Lincoln}
  \city{Lincoln, NE}
  \country{USA}}
\email{nathan.huynh@unl.edu}

\author{Hau Chan}
\affiliation{%
  \institution{University of Nebraska-Lincoln}
  \city{Lincoln, NE}
  \country{USA}}
\email{hchan3@unl.edu}


\begin{abstract}
Evolutionary code search frameworks driven by large language models (LLMs) can now automate the design of strong heuristics for hard optimization problems such as vehicle routing problems, but every one of them still requires human experts to first decide how the problem will be represented and to write the surrounding program that the search then improves.
Such decision is problem-specific and demanding for human experts, yet under-specified in reported results.
In this paper, we show that this bottleneck can be substantially mitigated via \emph{problem transformation}, where an LLM first proposes mappings from the problem at hand into several related problems it has good knowledge about, then an evolutionary code search evolves heuristics for these problems at once so that algorithmic ideas migrate between them.
On seven combinatorial optimization problems, including the last-mile delivery routing problem on real operational data, heuristics produced this way are mostly competitive with or better than those designed within handcrafted program skeletons while requiring roughly $6\times$ fewer lines of human-written scaffolding, though the most constrained one still needs handcrafted repair.
Ultimately, our approach provides an intuitive means to reduce the manual efforts involved in the ever-growing field of evolutionary code search.
\end{abstract}

\ccsdesc[500]{Computing methodologies~Search methodologies}
\ccsdesc[300]{Computing methodologies~Machine learning}
\ccsdesc[300]{Theory of computation~Discrete optimization}

\keywords{automatic heuristic design, combinatorial optimization, large language models, evolutionary computation, problem transformation}


\maketitle

\section{Introduction}\label{sec:intro}

In recent years, evolutionary code search driven by large language models (LLMs) has shown remarkable success in automating the design of high-quality heuristics for NP-hard combinatorial optimization problems, which traditionally requires extensive domain knowledge and manual efforts from human experts \citep{liu2026systematic}.
Such heuristics serve repeated, latency-bound decisions on web platforms, e.g., last-mile routing of e-commerce orders (Section \ref{subsec:lastmile}).
FunSearch \citep{romera2024mathematical} showed that by modeling the automatic heuristic design task as a search problem in the space of functions, where each function is a heuristic component represented by a (Python) program, LLMs can be integrated into an evolutionary framework to iteratively improve the quality of generated functions and subsequently surpass its human-designed counterparts in terms of optimization performance.
The success of FunSearch has lead to numerous follow-up works, most of which focus on sharpening the evolutionary machinery, e.g., a reflection step and multi-objective accommodation \citep{liu2024evolution,ye2024reevo,yao2024multi,zheng2025monte}.
Lately, AlphaEvolve \citep{novikov2025alphaevolve} pushed the same idea from evolving single functions to entire code files, i.e., from a single to multiple heuristic components, triggering a new family of LLM-based evolutionary code search frameworks \citep{sharma2025openevolve,lange2026shinkaevolve,assumpccao2025codeevolve}.

Nonetheless, despite the rapid advancement in this field, before any of its associated frameworks can be applied to a given optimization problem, human experts still need to manually define the search space.
For the function-level frameworks, i.e., FunSearch-style, this involves choosing an outer algorithm, such as ant colony optimization \citep{dorigo2007ant}, guided local search \citep{voudouris2010guided}, or (step-by-step) solution construction \citep{asani2023computation}, then implementing that algorithm for the problem at hand, deciding which subroutine inside it will be evolved, fixing that subroutine's function signature, and writing the evaluator and the LLM prompts respecting all of the above.
The more capable file-level frameworks, i.e., AlphaEvolve-style, actually increase this burden.
In fact, they require from users a self-contained initial program that already solves the problem \citep{novikov2025alphaevolve}, which is strictly more demanding than what required from FunSearch-style frameworks (see Table \ref{tab:spec-comparison} for a better picture).


\begin{table}[t]
    \centering
    \footnotesize
    \caption{Comparison of manual efforts required for representative evolutionary code search frameworks.
    \textbf{H} and \textbf{A} respectively denote ``handcrafted'' and ``automatic'' (i.e., generated by LLMs).
    The columns are formally defined in Section~\ref{subsec:spec}.
    The last column reports the median number of non-comment scaffold lines over the seven problems of Section \ref{sec:exp}, counted in accordance with Section \ref{subsec:effort}.}
    \begin{tabular}{l|ccccc|r}
        \thickhline
        Method & $\varsigma^{\mathrm{rep}}$ & $\varsigma^{\mathrm{skel}}$ & $\varsigma^{\mathrm{sig}}$ & $\varsigma^{\mathrm{eval}}$ & $\varsigma^{\mathrm{ppt}}$ & Lines \\\hline
        FunSearch \citeyear{romera2024mathematical} & H & H & H & H & H & 165 \\
        EoH \citeyear{liu2024evolution} & H & H & H & H & H & 210 \\
        ReEvo \citeyear{ye2024reevo} & H & H & H & H & H & 238 \\
        MCTS-AHD \citeyear{zheng2025monte} & H & H & H & H & H & 226 \\
        OpenEvolve \citeyear{sharma2025openevolve} & H & H & H & H & H & 297 \\\hline
        \textbf{ReFrame} (ours) & A & A & A & H & A (partial) & \textbf{38} \\
        \thickhline
    \end{tabular}
    \label{tab:spec-comparison}
\end{table}

\paragraph{Our Approach.}
In this paper, we focus on automating such specification instead of empowering the search.
The observation we build on is that the outer algorithm a human picks can be formally treated as a pair of functions, one that converts an instance of the problem at hand into an instance of some other problem the outer algorithm is effective at \emph{a posteriori}, and one that turns a solution of that other problem back into a solution of the original problem.
Solution construction for the traveling salesman problem, for instance, transforms the problem into a sequence of next-node choices and reassembles a tour from them.
From this perspective, we argue that the specification can also be written by the same LLM that writes the heuristics.
The idea is to prompt the LLM to propose several candidate target problems in natural language, then implement the instance map and the solution map for each as code, followed by the function signature that a heuristic for the target problem must satisfy, the last of which is exactly what an evolutionary code search needs.
By this means, existing FunSearch-style and AlphaEvolve-style frameworks can operate with reduced manual efforts.

\paragraph{Contributions.} We summarize our contributions as follows:
\begin{itemize}[leftmargin=*]
    \item We identify the \emph{task specification}, i.e., the solution representation, program skeleton, and evaluation harness a human writes for each new problem, as the binding manual cost in LLM-driven heuristic design, and show that it has not necessarily reduced as evolutionary code search frameworks have progressed (Table \ref{tab:spec-comparison}).
    \item We introduce ReFrame, a framework that synthesizes this specification by transforming the problem at hand into related problems the LLM is knowledgeable about, and evolving heuristics for several such problems jointly so that algorithmic ideas migrate between them.
    The synthesized specification drives four different evolutionary searches from both families without modification.
    \item We evaluate on seven optimization problems, including a last-mile delivery routing problem built on real operational data, and characterize where the approach holds and where it fails.
\end{itemize}

\paragraph{Outline.}
Section \ref{sec:related} discusses related work.
Section \ref{sec:prelim} provides the preliminaries.
Section \ref{sec:reframe} describes our proposed framework.
Sections \ref{sec:exp} and \ref{sec:analysis} evaluate its efficacy and analyze when it fails.
Sections \ref{sec:limit} and \ref{sec:conclu} conclude our work.
The appendices mentioned throughout the paper can be found in the supplementary material.

\section{Related Work}\label{sec:related}

\paragraph{Automatic Heuristic Design.}
Hyper-heuristics \citep{burke2013hyper,pillay2018hyper} generalize heuristic design by searching over heuristics rather than over solutions, either selecting from a predefined set \citep{drake2020recent} or composing new ones from simpler components \citep{duflo2019gp,zhao2023automated}.
Genetic programming \citep{langdon2013foundations} is the dominant generative variant and is valued for producing interpretable heuristics \citep{mei2022explainable}.
The bottleneck is that human experts must supply the primitive set and the permissible operators that define what can be built, which is a difficult task in practice and is itself a form of the specification burden our paper targets \citep{o2010open,pillay2018hyper}.

\paragraph{Evolutionary Code Search with LLMs.}
Advances in LLMs have enabled new approaches for automatic heuristic design \citep{liu2026systematic}.
Since standalone LLMs with prompt engineering are generally incapable of producing novel algorithmic ideas beyond their encoded knowledge \citep{mahowald2024dissociating}, most active research in this area focuses on integrating LLMs into an evolutionary code search to iteratively refine a set of heuristics.
The idea involves iteratively improving a set of candidate heuristics through score-based selection (i.e., identifying the ``fittest'' candidate heuristics subject to a so-called fitness function) and stochastic variation operators (e.g., crossover and mutation among the fittest candidate heuristics as inspired by biological evolution).
Representative frameworks in this field include FunSearch \citep{romera2024mathematical}, EoH \citep{liu2024evolution}, ReEvo \citep{ye2024reevo}, MEoH \citep{yao2024multi}, and MCTS-AHD \citep{zheng2025monte}.
All of these frameworks, at its core, evolve one function inside an outer algorithm that a human implements, and \citet{liu2026systematic,liu2024large} note this dependence as the field's standing limitation.
AlphaEvolve \citep{novikov2025alphaevolve} scales the evolutionary code search to entire files by having an LLM ensemble emit diffs against a user-supplied program in which regions to evolve are marked, scored by a user-supplied \texttt{evaluate} function.
Open reimplementations, such as OpenEvolve \citep{sharma2025openevolve}, ShinkaEvolve \citep{lange2026shinkaevolve}, and CodeEvolve \citep{assumpccao2025codeevolve}, reproduce and in places improve on the reported results.
These frameworks widen what the search may edit, but they do not narrow what the user must first write, hence the same limitation as FunSearch-style frameworks remains.

\paragraph{Neural Combinatorial Optimization.}
A parallel line of work in automatic heuristic design employs neural networks to search for the optimal parameter settings within a parameterized heuristic space \citep{bengio2021machine, yang2023survey}. 
Compared to LLM-based evolutionary code search, they are resource-intensive \citep{kwon2020pomo}, hard to implement \citep{zheng2024udc}, and may yield subpar results in various experimental settings \citep{liu2024evolution,ye2024reevo,zheng2025monte}. 


\section{Preliminaries}\label{sec:prelim}

\subsection{Optimization Problems and Heuristics}\label{subsec:cop}

We first formally define a combinatorial optimization problem. 

\begin{definition}\label{def:cop}
A combinatorial optimization problem $A$ is a triplet $(\mathcal{X}_A,\mathcal{S}_A,q_A)$, where $\mathcal{X}_A$ is a set of problem instances, $\mathcal{S}_A$ assigns to each instance $x\in\mathcal{X}_A$ a finite set $\mathcal{S}_A(x)$ of \emph{feasible solutions}, and $q_A:\{(x,y): x\in\mathcal{X}_A,\, y\in\mathcal{S}_A(x)\}\to\mathbb{R}$ is an objective.
We refer to solving $A$ as finding a feasible solution $y\in\mathcal{S}_A(x)$ for a given instance $x$ such that $q_A(x,y)$ is as large as possible.
\end{definition}

For example, in the capacitated vehicle routing problem, $\mathcal{S}_A(x)$ contains exactly those sets of routes that start and end at the depot, visit every customer once, and respect the vehicle capacity (Appendix~A gives $\mathcal{S}_A$ and $q_A$ for every problem we evaluate on).
Minimization problems are handled by negating the objective, so that larger is always better.
A \emph{heuristic} for $A$ is a program $h$ with $h(x)\in\mathcal{S}_A(x)$, and given a distribution $\mathcal{D}$ over $\mathcal{X}_A$ and a space $H$ of heuristics an LLM might write, the search task is
\begin{equation}\label{eq:prob-def}
    h^{*}\in\argmax_{h\in H}\ \mathbb{E}_{x\sim\mathcal{D}}\big[q_A(x,h(x))\big],
\end{equation}
which in practice is approximated on a finite training set of $n_{\mathrm{tr}}$ instances drawn from $\mathcal{D}$. 

\subsection{The Task Specification}\label{subsec:spec}

Unfortunately, existing evolutionary code search frameworks do not search for $h$ directly as in Equation~\ref{eq:prob-def}, but rather for a program $h'$ that is a heuristic component inside a larger program supplied by human experts, so that the search task is instead
\begin{equation}\label{eq:prob-def-real}
    h^{*}\in\argmax_{h'\in H_B}\ \mathbb{E}_{x\sim\mathcal{D}}\Big[q_A\big(x,\,g(h'(f(x)))\big)\Big],
\end{equation}
where $f$ maps an instance of $A$ to the input the evolved component consumes, and $g$ assembles the component's output back into a solution of $A$.
For solution construction on the traveling salesman problem, for example, $f(x)$ presents the distance matrix together with the current partial tour and $g$ concatenates the chosen nodes, whereas for ant colony optimization, $g$ runs the sampling loop over the produced heuristic matrix.
The pair $(f,g)$ is normally not discussed in the methodologies of existing frameworks, yet building it is the bulk of the manual work.
Concretely, as summarized in Table~\ref{tab:spec-comparison}, applying an evolutionary code search to a problem $A$ requires human experts to supply a \emph{task specification} $\Sigma_A=(\varsigma^{\mathrm{rep}},\varsigma^{\mathrm{skel}},\varsigma^{\mathrm{sig}},\varsigma^{\mathrm{eval}},\varsigma^{\mathrm{ppt}})$, consisting of a solution representation $\varsigma^{\mathrm{rep}}$ that fixes what $h'$ consumes and produces, a program skeleton $\varsigma^{\mathrm{skel}}$ implementing $f$ and $g$ around the evolved component, the function signature $\varsigma^{\mathrm{sig}}$ that component must satisfy, an evaluator $\varsigma^{\mathrm{eval}}$ that scores a candidate component on training instances, and prompt fragments $\varsigma^{\mathrm{ppt}}$ respectively describing $A$, the heuristic component, and the function signature to the LLM.

\subsection{Problem Transformation}\label{subsec:transformation}

We observe that $(f,g)$ can be formalized as follows.

\begin{definition}\label{def:transformation}
Let $A$ and $B$ be combinatorial optimization problems.
A \emph{transformation} from $A$ to $B$ is a pair of programs $(f,g)$ with $f:\mathcal{X}_A\to\mathcal{X}_B$ mapping instances from $A$ to $B$ and $g$ mapping solutions of $B$ back to $A$.
The transformation is \emph{valid} if
\begin{equation}\label{eq:validity}
    \forall x\in\mathcal{X}_A,\ \forall y'\in\mathcal{S}_B(f(x)):\quad g(y')\in\mathcal{S}_A(x),
\end{equation}
that is, if every feasible solution of the transformed instance maps back to a feasible solution of the original instance.
\end{definition}

This resembles the notion of \emph{reduction} \citep{crescenzi1997short} in complexity theory but without any approximation-ratio guarantees.
Motivated by this observation, our work investigates whether LLMs can choose the target problem $B$ and associated $(f,g)$ by itself, circumventing the need to fully construct $\Sigma_A$.

\paragraph{Validity Check.}
Equation~\ref{eq:validity} quantifies over all instances and all feasible solutions of the transformed instance, which is unverifiable for finite $n_{\mathrm{tr}}$.
A transformation is therefore retained if, for each of the $n_{\mathrm{tr}}$ training instances, the solution it returns passes a feasibility predicate for $A$ written by hand once per problem, and we then report, on held-out instances, the fraction on which the returned solution is feasible.
We call this the transformation's \emph{empirical validity rate} and treat any value below $100\%$ as a failure of the method rather than as noise, a rate that Section \ref{sec:analysis} shows falling from $99\%$ to $23\%$ as constraints are added.
Where a hard guarantee is required, the evaluator can be wrapped in a repair step or in a constructive skeleton that enforces feasibility, at the cost of reintroducing part of $\Sigma_A$, and we measure that trade-off in Section \ref{subsec:lastmile}.

\paragraph{What Remains Handcrafted.}\label{subsec:residual}
Two components of $\Sigma_A$ remain handcrafted for every new problem, and the claims in this paper are stated net of them.
The first is the prose description of $A$ that seeds the transformation prompt, which must be written carefully, since Section \ref{subsec:retrieval} shows that rephrasing it into unfamiliar terminology degrades the transformations the model proposes.
The second is the feasibility predicate used by the evaluator, which is what makes the empirical check above possible at all and is not derivable from the prose description.
Both are properties of the problem rather than of any search algorithm, so they are written once and reused across every search we run, whereas the skeleton, signature, and prompt fragments in Table \ref{tab:spec-comparison} must be rewritten whenever either the problem or the outer algorithm changes.
Together they came to a median of $38$ non-comment lines per problem, against $226$ for the median baseline (Section \ref{subsec:effort}), which we report as a reduction in scaffold lines and not as the elimination of specification effort.

\section{The ReFrame Framework}\label{sec:reframe}

\begin{figure*}[t]
    \centering
    \includegraphics[width=0.98\textwidth]{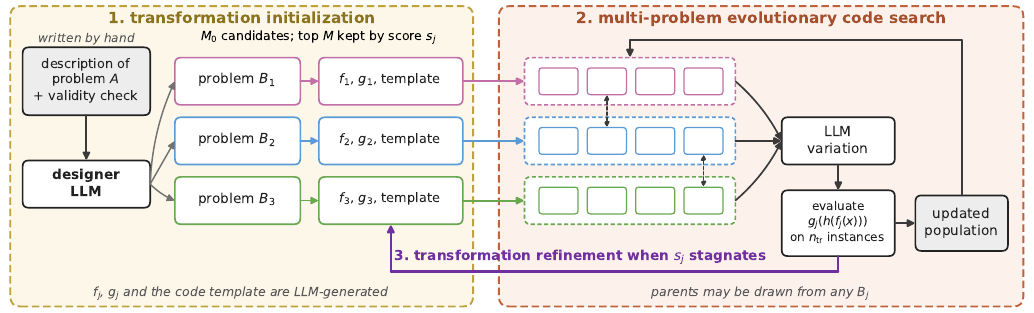}
    \caption{ReFrame.
    Stage 1 turns a prose description of the problem at hand into $M_0$ candidate transformations, each consisting of a target problem, an instance map, a solution map, and the code template that fixes the signature of the heuristic to be evolved, of which the $M$ best by score are retained.
    Stage 2 evolves heuristics for all $M$ target problems in one population, drawing parents across target problems so that an idea found under one representation can be applied under another.
    Stage 3 rewrites a transformation whose score has stopped improving, keeping the rewrite only if the score improves.
    Everything inside the two panels is generated by the LLM, and the two gray boxes on the left are the handcrafted input.}
    \label{fig:framework}
\end{figure*}

As illustrated in Figure \ref{fig:framework}, ReFrame takes as input the prose description of $A$ and its feasibility predicate, and returns a heuristic for $A$.
Between the two it synthesizes and then repairs a set of task specifications, and evolves heuristics against all of them at once.
Table \ref{tab:notation} lists the notation used in this section.
The exact prompts used within ReFrame can be found in Appendix~C.1.

\begin{table}[h]
    \centering
    \footnotesize
    \caption{Notation.}
    \begin{tabular}{ll}
        \thickhline
        Symbol & Meaning \\\hline
        $A$, $B_j$ & the problem at hand; the $j$-th target problem \\
        $(f_j,g_j)$ & instance map and solution map of the $j$-th transformation \\
        $r_j$ & the $j$-th transformation: $B_j$, $(f_j,g_j)$, and its code template \\
        $M_0$, $M$ & candidate transformations generated; transformations retained \\
        $\mu$ & size of the heuristic population $P$ \\
        $\mu_j$ & offspring generated for $B_j$ in one generation, $\sum_j\mu_j=\mu$ \\
        $\ell$ & number of top heuristics averaged into a transformation's score \\
        $s_j$ & score of transformation $r_j$ (Equation~\ref{eq:score}) \\
        $n_{\mathrm{tr}}$ & number of training instances drawn from $\mathcal{D}$ \\
        $T$ & generations without improvement in $s_j$ before $r_j$ is refined \\
        \thickhline
    \end{tabular}
    \label{tab:notation}
\end{table}

\subsection{Transformation Initialization}\label{subsec:init}

\paragraph{What a Transformation Is.}
A transformation $r_j$ bundles four things: a description of the target problem $B_j$ in one or two sentences of natural language; the code implementing the instance map $f_j$ and the solution map $g_j$; the code template for $B_j$, which fixes the signature $\varsigma^{\mathrm{sig}}$ that a heuristic for $B_j$ must satisfy and is what a human would otherwise write by hand; and a score $s_j$, defined below, that drives selection and stagnation detection (see Figure~S7 for a complete example).

\paragraph{Generation.}
Given the description of $A$, the LLM is first asked for $M_0\ge M$ candidate target problems, described in prose and without reference to $A$, then asked for each candidate, in a separate call, to implement $(f_j,g_j)$ against the transformation template, and in a third call to emit the code template for $B_j$ from those two functions.
Splitting the requests rather than asking for everything at once improves the result and reduces hallucinated references between the parts \citep{huang2025survey}, and Appendix~C.1 gives the prompts and the comparison.

\paragraph{Scoring and Selection.}
For each candidate we initialize a population $P_j$ of $\lceil\mu/M\rceil$ heuristics by prompting the LLM with $B_j$'s description and code template, and a heuristic $h^{(j)}$ for $B_j$ is scored on $A$, not on $B_j$:
\begin{equation}\label{eq:fitness}
    Q\big(h^{(j)}\big)=\frac{1}{n_{\mathrm{tr}}}\sum_{k=1}^{n_{\mathrm{tr}}} q_A\Big(x^{(k)},\ g_j\big(h^{(j)}(f_j(x^{(k)}))\big)\Big),
\end{equation}
with $Q(h^{(j)})=-\infty$ if the returned solution fails the feasibility predicate on any training instance.
The score of the transformation itself is the mean of its best $\ell$ heuristics,
\begin{equation}\label{eq:score}
    s_j=\frac{1}{\ell}\sum_{h\in \mathrm{top}_\ell(P_j)}Q(h),
\end{equation}
which is finite exactly when at least $\ell$ of its heuristics are feasible everywhere on the training set.
The $M$ candidates with the highest $s_j$ are retained and the rest discarded, giving the initial population $P=\bigcup_{j=1}^{M}P_j$ of at least $\mu$ heuristics, and the fraction of the $M_0$ candidates with a finite score is the initialization-time validity rate we report in Section \ref{subsec:chosen}.

\subsection{Multi-Problem Evolutionary Code Search}\label{subsec:multiproblem}

With $M$ transformations retained, the search proceeds as in any evolutionary code search, i.e., select parents, prompt the LLM to apply a variation operator, evaluate the offspring by Equation~\ref{eq:fitness}, and keep the fittest $\mu$ heuristics.
The one change we make is in parent selection.
Because the population now contains heuristics written against $M$ different signatures, the natural implementation would partition it and evolve each slice separately.
We instead do the opposite.
A parent may be any heuristic in $P$, regardless of which target problem it was written for, so a heuristic $h^{(j)}$ for $B_j$ can serve as the algorithmic reference from which an offspring for $B_{j'}$, $j'\neq j$, is generated.
This prevents a single transformation from monopolizing the population, and more importantly it lets an algorithmic idea developed under one representation be tried under another, which is where the gains in our experimental results (particularly Table~S5 in Appendix~C.2) come from.
Figure~S6 in Appendix~C.3 traces one such transfer, in which a $2$-opt edge exchange written for one target problem is carried by the mutation operator into a population where nothing had suggested it, and the offspring is better than any of its siblings.

\paragraph{Allocating Offspring Across Target Problems.}
Each variation operator produces $\mu$ offspring per generation, split across the $M$ target problems as $\mu_1,\dots,\mu_M$ with $\sum_j\mu_j=\mu$.
The split is drawn by sampling $\mu$ times from $\{r_1,\dots,r_M\}$ with probability $p_j\propto 1/|s_j|$ when $q_A<0$ (minimization, e.g., tour length) and $p_j\propto s_j$ when $q_A\ge 0$ (maximization, e.g., packed value), mirroring how EoH selects parent heuristics \citep{liu2024evolution}.
Better transformations therefore receive more of the budget without any of them being cut off, and the total number of LLM calls per generation is unchanged from the single-specification case.
We describe the procedure using EoH as the underlying search because of its simplicity, and Section \ref{subsec:ablations} runs the same construction on three others.

\subsection{Transformation Refinement}\label{subsec:refine}

A transformation that was implemented poorly will still attract budget if its score happens to be competitive early, and the search around it then converges to a local optimum from which no variation of the heuristic can escape.
That is, the limitation is in $(f_j,g_j)$ and not in $h^{(j)}$, so ReFrame rewrites transformations rather than only heuristics.
When $s_j$ has not improved for $T$ consecutive generations, the LLM is prompted with the descriptions of $A$ and $B_j$ and the current $(f_j,g_j)$ and code template, and asked for a revised implementation.
The heuristics associated with $r_j$ are then re-evaluated through the new maps, which updates $s_j$, and the rewrite is kept only if $s_j$ improves.
This is the single most load-bearing component of the framework in our ablations (Section \ref{subsec:ablations}), and it costs one LLM call plus $\mu_j$ re-evaluations each time it fires.

To make the three stages concrete, one run on the capacitated vehicle routing problem turned a five-sentence description and a feasibility predicate into a target problem described as ``partitioning $n$ weighted points into groups whose total weight does not exceed a bound, then ordering the points within each group to minimize travel'', together with the two maps and the code template a human would otherwise have written (Appendix~C.3).

\section{Experiments}\label{sec:exp}

\subsection{Setup}\label{subsec:setup}

\paragraph{Problems and Data.}
Six of our seven problems follow the setup of MCTS-AHD \citep{zheng2025monte}, one of the most capable published evolutionary code search frameworks for the respective family, using their public instance sets for training and testing: the traveling salesman problem (TSP), the capacitated vehicle routing problem (CVRP), the $0/1$ knapsack problem (KP), the multiple knapsack problem (MKP), and the offline and online bin packing problems (BPP and OBPP), all defined in Appendix~A.
The seventh is a last-mile delivery routing problem with time windows (Section \ref{subsec:lastmile}), and following prior work we also report on TSPLib \citep{reinelt1991tsplib}.
Training uses $n_{\mathrm{tr}}=64$ instances per problem except for OBPP, where we follow \citet{zheng2025monte} in using a 5k-item stream, and out-of-distribution settings use sizes never seen during the search.

\paragraph{Framework Settings.}
Unless stated otherwise we use $M_0=10$ candidate transformations, $M=3$ retained, $\ell=3$, $\mu=20$, $T=3$, and GPT-4o-mini at temperature $1$ as the designer LLM for both transformations and heuristics.
The underlying search is EoH \citep{liu2024evolution} with two variation operators, one crossover and one mutation, rather than the five of the original work (see Appendix~C.1).
Each heuristic is given at most $60$ seconds on the training set, and the whole three-stage procedure is capped at $1{,}000$ fitness evaluations, which is at or below the budget used in the works we compare against \citep{liu2024evolution,zheng2025monte}.

\paragraph{Baselines and Reference Points.}
We compare against three groups of methods: handcrafted heuristics and approximation algorithms \citep{rosenkrantz1977analysis,brecklinghaus2015approximation,christofides1976worst}; learned solvers, namely POMO \citep{kwon2020pomo} and DeepACO \citep{ye2023deepaco}; and evolutionary code searches, namely FunSearch, EoH, ReEvo, and MCTS-AHD, each inside the outer algorithm its authors used.
We add LKH-3 \citep{helsgaun2017extension}, OR-Tools routing \citep{ortools_routing}, and HGS \citep{vidal2022hybrid} as reference points rather than as competitors, since the goal of this work is to automate heuristic \emph{design} and not to outperform solvers that have absorbed decades of engineering.
Those rows exist to make the remaining distance visible, which Section \ref{subsec:cost} pairs with the time each method needs.

\paragraph{Reporting.}
Optimality gaps are reported against exact references and never against a heuristic, namely Concorde \citep{applegate2006traveling} for TSP, exact dynamic programming for KP, CP-SAT for MKP, and the standard bin lower bound for BPP and OBPP, whereas for CVRP and the last-mile task we report objective values directly, since no proven optima are available at these sizes.
Every number we report for our own method is the mean over five independent runs, with the standard deviation across runs, whereas baseline numbers taken from the literature are reproduced as published and marked as such in the captions.

\subsection{Main Results}\label{subsec:res}

Published evolutionary code searches cannot be run without an outer algorithm, so we compare against them inside the two the literature uses, namely step-by-step solution construction and ant colony optimization, whereas our method uses neither.

\paragraph{Against Solution Construction.}
As shown in Table \ref{tab:main-sbs}, on TSP, heuristics designed against synthesized specifications beat every search that used the handcrafted construction skeleton, by $7.4\%$ at $n=50$ and $7.9\%$ at $n=200$ against the strongest of them, and at $n=200$ the result is also better than POMO, which was trained for the problem.
On KP every method is within $0.2\%$ of optimal and ours is second best, within $0.05\%$ of MCTS-AHD.
On OBPP we are clearly worse than the specialized searches, and we attribute this to the fact that an online packing decision must be made per arriving item with no view of the future, which leaves only a small space of target problems that preserve the restriction, so that the transformations proposed largely rediscover scoring rules already encoded by the construction skeleton.
Section \ref{subsec:ablations} shows part of this gap closing with a stronger designer LLM.

\begin{table}[h]
    \centering
    \footnotesize
    \caption{Comparison against searches operating inside a handcrafted solution-construction skeleton: (top) TSP and KP, (bottom) OBPP.
    Searches marked with an asterisk consume that skeleton, whereas ReFrame does not.
    Baseline numbers are as reported by \citet{zheng2025monte}, and our rows are mean\,$\pm$\,std over five runs.
    $n$ is the number of nodes for TSP and the number of items for KP and OBPP; $W$ is the knapsack capacity for KP and the bin size for OBPP.
    Underlined settings are in-distribution, the rest out-of-distribution.
    The best method using GPT-4o-mini is shaded and the overall best is bold.}
    \begin{tabular}{l|ccc|ccc}
        \thickhline
        \multirow{3}{*}{\backslashbox{Method}{\strut Problem\\ setting}} & \multicolumn{3}{c|}{TSP (Obj. $\downarrow$)} & \multicolumn{3}{c}{KP (Obj. $\uparrow$)} \\\cline{2-7}
        & \underline{$n$=50} & $n$=100 & $n$=200 & \makecell{\underline{$n$=50}\\ \underline{$W$=12.5}} & \makecell{$n$=100\\$W$=25} & \makecell{$n$=200\\$W$=25} \\\hline
        Optimal (Concorde / exact DP) & 5.688 & 7.756 & 10.717 & 20.031 & 40.278 & 57.447 \\\hline
        Greedy \citeyear{rosenkrantz1977analysis} & 6.959 & 9.706 & 13.461 & 19.985 & 40.225 & 57.395 \\
        POMO \citeyear{kwon2020pomo} & \textbf{5.697} & \textbf{8.001} & 12.897 & 19.612 & 39.676 & 57.271 \\\hline
        FunSearch* & 6.357 & 8.850 & 12.372 & 19.988 & 40.227 & 57.398 \\
        EoH* & 6.394 & 8.894 & 12.437 & 19.993 & 40.231 & 57.399 \\
        MCTS-AHD* & 6.225 & 8.684 & 12.120 & \cellcolor{gray!25}\textbf{20.015} & \cellcolor{gray!25}\textbf{40.252} & \cellcolor{gray!25}\textbf{57.423} \\\hline
        ReFrame & \cellcolor{gray!25}5.767 & \cellcolor{gray!25}8.006 & \cellcolor{gray!25}\textbf{11.164} & 20.006 & 40.248 & 57.416 \\
        & \tiny$\pm$0.019 & \tiny$\pm$0.031 & \tiny$\pm$0.062 & \tiny$\pm$0.004 & \tiny$\pm$0.006 & \tiny$\pm$0.008 \\
        \thickhline
    \end{tabular}\\
    \medskip
    \begin{tabular}{l|cccccc}
        \thickhline
        \multicolumn{7}{c}{OBPP (\% gap to the bin lower bound $\downarrow$)} \\\hline
        \multicolumn{1}{c|}{$n$} & \underline{1k} & \underline{1k} & \underline{5k} & \underline{5k} & 10k & 10k \\
        \multicolumn{1}{c|}{$W$} & \underline{100} & \underline{500} & \underline{100} & \underline{500} & 100 & 500 \\\hline
        Best Fit & 4.77 & \textbf{0.25} & 4.31 & 0.55 & 4.05 & 0.47 \\
        First Fit & 5.02 & \textbf{0.25} & 4.65 & 0.55 & 4.36 & 0.50 \\\hline
        FunSearch* & \textbf{2.45} & 0.66 & 1.30 & \textbf{0.25} & 1.05 & \textbf{0.21} \\
        EoH* & 2.69 & \textbf{0.25} & 1.63 & 0.53 & 1.47 & 0.45 \\
        ReEvo* & 3.94 & 0.50 & 2.72 & 0.40 & 2.39 & 0.31 \\
        MCTS-AHD* & \textbf{2.45} & 0.50 & \textbf{1.06} & 0.32 & \textbf{0.74} & 0.26 \\\hline
        ReFrame & 3.78 & 0.99 & 2.82 & 0.55 & 2.61 & 0.40 \\
        & \tiny$\pm$0.13 & \tiny$\pm$0.10 & \tiny$\pm$0.16 & \tiny$\pm$0.07 & \tiny$\pm$0.18 & \tiny$\pm$0.06 \\
        \thickhline
    \end{tabular}
    \label{tab:main-sbs}
\end{table}

\begin{table*}[t]
    \centering
    \footnotesize
    \caption{Comparison against searches operating inside a handcrafted ant colony optimization skeleton, which the searches marked with an asterisk consume and ReFrame does not.
    Baseline numbers are as reported by \citet{zheng2025monte}, our rows are mean\,$\pm$\,std over five runs, and the two reference rows are dedicated solvers given $60$ seconds per instance.
    $n$: nodes for TSP and CVRP, items for MKP and BPP; $C$: vehicle capacity; $m$: number of knapsacks; $W$: bin size.
    Underlined settings are in-distribution, the best method using GPT-4o-mini is shaded, and the overall best among the designed heuristics is bold.}
    \begin{tabular}{l|cc|cc|cc|cc}
        \thickhline
        \multirow{3}{*}{\backslashbox{Method}{\strut Problem\\ setting}} & \multicolumn{2}{c|}{TSP (Obj. $\downarrow$)} & \multicolumn{2}{c|}{CVRP (Obj. $\downarrow$)} & \multicolumn{2}{c|}{MKP (Obj. $\uparrow$)} & \multicolumn{2}{c}{BPP (Obj. $\downarrow$)} \\\cline{2-9}
        & \underline{$n$=50} & $n$=100 & \makecell{\underline{$n$=50}\\ \underline{$C$=50}} & \makecell{$n$=100\\$C$=50} & \makecell{\underline{$n$=100}\\\underline{$m$=5}} & \makecell{$n$=200\\$m$=5} & \makecell{\underline{$n$=500}\\\underline{$W$=150}} & \makecell{$n$=1,000\\$W$=150} \\\hline
        \emph{Ref.:} LKH-3 \citeyear{helsgaun2017extension} & 5.688 & 7.756 & 8.145 & 13.352 & -- & -- & -- & -- \\
        \emph{Ref.:} OR-Tools / CP-SAT & 5.702 & 7.803 & 8.314 & 13.948 & 23.418 & 42.947 & 201.0 & 400.4 \\\hline
        ACO \citeyear{dorigo2007ant} & 5.992 & 8.948 & 11.355 & 18.778 & 22.738 & 40.672 & 208.828 & 417.938 \\
        DeepACO \citeyear{ye2023deepaco} & 5.842 & 8.282 & \textbf{8.888} & \textbf{14.932} & 23.093 & 41.988 & \textbf{203.125} & \textbf{405.172} \\\hline
        EoH* & 5.828 & 8.263 & 9.359 & \cellcolor{gray!25}15.681 & 23.139 & 41.994 & 204.646 & 408.599 \\
        ReEvo* & 5.856 & 8.340 & 9.327 & 16.092 & 23.245 & 42.416 & 206.693 & 413.510 \\
        MCTS-AHD* & \cellcolor{gray!25}\textbf{5.801} & 8.179 & \cellcolor{gray!25}9.286 & 15.782 & \cellcolor{gray!25}\textbf{23.269} & 42.498 & 204.094 & 407.323 \\\hline
        ReFrame & 5.819 & \cellcolor{gray!25}\textbf{8.039} & 9.826 & 15.726 & 23.164 & \cellcolor{gray!25}\textbf{42.682} & \cellcolor{gray!25}203.344 & \cellcolor{gray!25}405.359 \\
        & \tiny$\pm$0.021 & \tiny$\pm$0.034 & \tiny$\pm$0.115 & \tiny$\pm$0.162 & \tiny$\pm$0.048 & \tiny$\pm$0.093 & \tiny$\pm$0.421 & \tiny$\pm$0.685 \\
        \thickhline
    \end{tabular}
    \label{tab:main-aco}
\end{table*}

\paragraph{Against Ant Colony Optimization.}
Table \ref{tab:main-aco} repeats the comparison against the stronger skeleton.
Without any skeleton of its own, ReFrame is ahead of every asterisked search on both out-of-distribution settings for TSP, MKP and BPP, and competitive in-distribution, where it is within $0.5\%$ of the best asterisked search.
CVRP is the exception, where we are $5.8\%$ behind the best asterisked search in-distribution, and we attribute this to the lack of domain knowledge from GPT-4o-mini, which proposes weak partitionings of the customer set.
Section \ref{subsec:ablations} shows that this gap disappears with a reasoning model.
The reference rows put all of this in perspective, since LKH-3 is $3.6\%$ better than the best designed heuristic on TSP at $n=100$ and $15\%$ better on CVRP, which is what one should expect from a solver written for the problem and which Section \ref{subsec:cost} balances against the time it takes.

\paragraph{Real-World Instances.}
On the $21$ TSPLib instances, whose geometry is unlike the uniform-random training data, the heuristic designed here is best on every instance and holds a single-digit optimality gap above $1{,}500$ nodes, where the asterisked searches were not run (Table~S7).
We read this as a generalization result about that particular heuristic rather than about the framework.

\subsection{What It Costs}\label{subsec:cost}

Solution quality alone does not decide anything, since a reader can always run an exact solver instead, so we report both the one-time expense of designing a heuristic and the per-instance expense of running it afterwards.
Designing a heuristic takes under $100$ minutes and under $\$0.15$ per problem with GPT-4o-mini, and $6$--$7\times$ that with o3-mini, because the search budget is capped at $1{,}000$ fitness evaluations.
The overhead over running the same search inside a handcrafted skeleton is transformation initialization and refinement, roughly a third of wall-clock and a tenth of LLM calls (Table~S8), and in return the skeleton itself no longer has to be written.

Against dedicated solvers on the two hardest routing settings, as shown in Table~\ref{tab:solver-ref}, the designed heuristics are one to three orders of magnitude faster per instance ($70$\,ms against $2.1$\,s for LKH-3 on TSP at $n=200$, and $0.21$\,s against $60$\,s for HGS on CVRP), and give up $2.5\%$ and $1.6\%$ of objective value respectively.
That trade is uninteresting when a handful of instances must be solved once, and it is the entire point when the same problem is solved thousands of times a day under a latency budget, the regime of Section \ref{subsec:lastmile}.

\begin{table}[h]
    \centering
    \footnotesize
    \caption{Quality against time on two hardest routing settings.
    Design time is a one-time cost, whereas inference time is per instance, and dedicated solvers are given the budgets shown.}
    \begin{tabular}{l|rr|rr}
        \thickhline
        \multirow{2}{*}{Method} & \multicolumn{2}{c|}{TSP, $n$=200} & \multicolumn{2}{c}{CVRP, $n$=100, $C$=50} \\\cline{2-5}
        & Obj. $\downarrow$ & Time/inst. & Obj. $\downarrow$ & Time/inst. \\\hline
        Concorde (exact) & \textbf{10.717} & 8.4\,s & -- & -- \\
        HGS \citeyear{vidal2022hybrid} & -- & -- & \textbf{13.302} & 60\,s \\
        LKH-3 \citeyear{helsgaun2017extension} & \textbf{10.717} & 2.1\,s & 13.352 & 60\,s \\
        OR-Tools \citeyear{ortools_routing} & 11.086 & 10.0\,s & 13.948 & 60\,s \\\hline
        EoH + ant colony opt. & 11.902 & 3.2\,s & 15.681 & 4.1\,s \\
        MCTS-AHD + construction & 12.120 & 0.06\,s & -- & -- \\\hline
        ReFrame (GPT-4o-mini) & 11.164 & 0.07\,s & 15.726 & 0.19\,s \\
        ReFrame (o3-mini) & 10.982 & 0.07\,s & 13.516 & 0.21\,s \\
        \thickhline
    \end{tabular}
    \label{tab:solver-ref}
\end{table}

\subsection{Last-Mile Delivery Routing}\label{subsec:lastmile}

The six problems above are the standard benchmark set for this literature and are all synthetic, so we add a seventh from an operational setting, namely routing delivery vehicles over customer stops with service times and delivery time windows to minimize total route duration.
We build $120$ instances from real routes driven in Austin and Boston in the 2021 Amazon Last Mile Routing Research Challenge data set \citep{merchan2024amazon}, with a mean of $147$ stops each, of which $64$ are used for the search and $56$ held out, and we add the Gehring--Homberger $200$-customer extension \citep{gehring1999parallel} of the Solomon benchmark \citep{solomon1987algorithms} as a synthetic counterpart.
Neither set was seen by any component of the framework during the six experiments above, and Appendix~C.4 gives both constructions.

We report total route duration and, separately, the fraction of held-out instances on which the returned solution is feasible, i.e., every stop is served within its window and no vehicle exceeds capacity (Table \ref{tab:lastmile}).
The two must be read together, feasibility first, and this is the first setting in the paper where they come apart.
With GPT-4o-mini as the designer, the synthesized transformations produce good routes when they produce feasible ones at all, but they are feasible on only $61.3\%$ of held-out real instances and $57.4\%$ of the synthetic ones, and a reasoning model raises this to $88.7\%$ and $86.2\%$ without fully resolving it (on average, about $22$ and then $6$ of the $56$ real routes fail).
Wrapping the evaluator in a handcrafted repair step, a constructive pass of roughly $30$ lines that inserts any window-violating stop at its cheapest feasible position, restores feasibility everywhere at a small cost in duration, and the designed heuristic is then level with OR-Tools on the real instances and within $2\%$ of it on the synthetic ones, while running more than $200\times$ faster per instance.
That repair step is exactly the kind of scaffold this paper set out to remove, and we report it as a limitation rather than as a component.
On a problem this tightly constrained, part of the specification has to come back, and Section \ref{sec:analysis} takes this apart.

\begin{table}[h]
    \centering
    \footnotesize
    \caption{Last-mile delivery routing.
    Feasibility is the fraction of held-out instances on which the returned solution serves every stop within its window without exceeding capacity.
    Duration is the mean total route duration in minutes over the instances a method solves feasibly, so rows are ranked feasibility-first, and below $100\%$ it is a conditional mean ($\dagger$) over a different subset, not comparable with the other rows.
    Solver rows are given $60$ seconds per instance.
    The asterisked rows use the constructive skeleton for routing with time windows of \citet{liu2024llm4ad}, which enforces feasibility by construction, and ``+ repair'' adds a handcrafted feasibility repair pass of roughly $30$ lines to our evaluator.}
    \begin{tabular}{l|rr|rr}
        \thickhline
        \multirow{2}{*}{Method} & \multicolumn{2}{c|}{Amazon last-mile} & \multicolumn{2}{c}{Gehring--Homberger} \\\cline{2-5}
        & Dur. $\downarrow$ & Feas. $\uparrow$ & Dur. $\downarrow$ & Feas. $\uparrow$ \\\hline
        OR-Tools \citeyear{ortools_routing} & 396.2 & 100\% & 4318.6 & 100\% \\
        LKH-3 \citeyear{helsgaun2017extension} & \textbf{388.4} & 100\% & \textbf{4102.4} & 100\% \\\hline
        EoH* + construction & 421.7 & 100\% & 4690.2 & 100\% \\
        MCTS-AHD* + construction & 417.3 & 100\% & 4612.8 & 100\% \\\hline
        ReFrame (GPT-4o-mini) & 438.9\rlap{$^\dagger$} & 61.3\% & 4884.1\rlap{$^\dagger$} & 57.4\% \\
        ReFrame (o3-mini) & 402.5\rlap{$^\dagger$} & 88.7\% & 4455.7\rlap{$^\dagger$} & 86.2\% \\
        ReFrame (o3-mini) + repair & 394.1 & 100\% & 4398.5 & 100\% \\
        \thickhline
    \end{tabular}
    \label{tab:lastmile}
\end{table}

\subsection{Ablations}\label{subsec:ablations}

We present four experiments isolating the components of ReFrame. 

\paragraph{Transformation Refinement, $M$, and $T$.}
As shown in Table~S9, with $T=3$, refinement fired between one and three times per run, and disabling it costs performance on every problem, with TSP falling from $5.819$ to $5.847$ and CVRP from $9.826$ to $10.218$ at $n=50$, behind the search the framework is built on.
Setting $M=1$ removes multi-problem search entirely and is clearly worse ($10.327$ on CVRP), but $M=3$ without cross-problem parents is worse still ($10.537$; Table~S5), so the gain comes from cross-problem parent selection, not from retaining more transformations, whereas $M=5$ and $M=7$ add little while multiplying the initialization cost (Table~S11).
The behavior of the patience parameter $T$ is less obvious, and we attribute this to the fact that rewriting a transformation after a single stagnant generation discards representations whose heuristic space had not yet been explored.

\paragraph{The Designer LLM\@.}
Everything above used GPT-4o-mini, a small general-purpose model known to be weak at algorithmic reasoning \citep{yang2025nondeterministic}, and the two places where our results were weakest (OBPP and CVRP) were both attributed to the transformations it proposed rather than to the search.
As shown in Table~S10, simply swapping in o3-mini as the designer LLM narrows both gaps, as OBPP improves in all settings, some of which even surpass the best baseline, and CVRP improves by $15.0\%$ in-distribution and $14.1\%$ out-of-distribution, moving from behind OR-Tools to within $1.6\%$ of HGS.
Since the search, the budget, and the evaluator are identical, the difference is attributable to the designer LLM, but as it writes both the transformations and the heuristics, this comparison alone does not isolate the transformation step; Section \ref{subsec:retrieval}, which changes only that step's input, shows that it matters on its own.

\paragraph{The Underlying Search.}
The claim in Section \ref{sec:intro} was that a synthesized specification is consumed by an evolutionary search unchanged, so we replace EoH with ReEvo, MEoH, and OpenEvolve \citep{sharma2025openevolve}, the most widely used open reimplementation of AlphaEvolve, and the integration required no change to either side.
We pass the generated code template as the initial program with \texttt{EVOLVE-BLOCK} markers around the heuristic body, and the fitness harness of Equation~\ref{eq:fitness} as the \texttt{evaluate} function that the LLM asks for, so that everything the LLM needs our transformation step already produced.
As shown in Table~\ref{tab:abl-eps}, replacing the handcrafted skeleton with a synthesized specification improves every one of the four searches on TSP, by $2.7\%$ for EoH at $n=100$ and by $3.1\%$ for OpenEvolve, but not on CVRP at $n=50$, where every synthesized row trails its skeleton-based counterpart.
OpenEvolve is the strongest search with and without our front-end, which is what ``orthogonal'' means operationally, since stronger search and synthesized specification compose and neither substitutes for the other.

\begin{table}[h]
    \centering
    \footnotesize
    \caption{The same synthesized specification driving four evolutionary searches.
    Rows marked with an asterisk run the search inside a handcrafted ant colony optimization skeleton, whereas rows marked ReFrame[$\cdot$] replace that skeleton with a synthesized specification and change nothing else.
    Mean over five runs except EoH* and ReEvo* (Appendix~C.1).
    Standard deviations are within 0.04 for TSP and 0.18 for CVRP.}
    \begin{tabular}{l|cc|cc}
        \thickhline
        \multirow{2}{*}{Search} & \multicolumn{2}{c|}{TSP (Obj. $\downarrow$)} & \multicolumn{2}{c}{CVRP (Obj. $\downarrow$)} \\\cline{2-5}
        & \underline{$n$=50} & $n$=100 & \underline{$n$=50} & $n$=100 \\\hline
        Best baseline & 5.801 & 8.179 & 9.286 & 15.681 \\\hline
        EoH* & 5.828 & 8.263 & 9.359 & 15.681 \\
        ReEvo* & 5.856 & 8.340 & 9.327 & 16.092 \\
        MEoH* & 5.812 & 8.201 & 9.294 & 15.907 \\
        OpenEvolve* & 5.794 & 8.106 & \textbf{9.238} & 15.462 \\\hline
        ReFrame[EoH] & 5.819 & 8.039 & 9.826 & 15.726 \\
        ReFrame[ReEvo] & 5.835 & 8.251 & 9.918 & 15.994 \\
        ReFrame[MEoH] & 5.730 & 7.883 & 9.702 & 15.503 \\
        ReFrame[OpenEvolve] & \textbf{5.712} & \textbf{7.854} & 9.564 & \textbf{15.281} \\
        \thickhline
    \end{tabular}
    \label{tab:abl-eps}
\end{table}


\subsection{Specification Effort}\label{subsec:effort}

Table \ref{tab:spec-comparison} claims a roughly sixfold reduction in handcrafted scaffold, and that number needs a protocol.
We count non-comment, non-blank lines of Python that a competent user must write to point a framework at one new problem, excluding for all methods anything reusable across problems, i.e., the framework's own source, imports, instance loading, and the objective function.
For the asterisked searches this is the outer algorithm's implementation for that problem, the evolved component's signature, the evaluator, and the prompt fragments (Appendix~B shows the ant colony optimization scaffolds for TSP and CVRP side by side), for AlphaEvolve-family frameworks it is the complete initial program, the marked regions, and the \texttt{evaluate} function, and for our framework it is the prose description of $A$, i.e., the problem statement and the template fragments of Tables~S3 and~S4, and the feasibility predicate.
Counts were taken from the released implementations where they exist and from our own ports otherwise, giving $38$ lines against $226$ for the median baseline, a factor of $5.9$.
Three caveats apply.
Line counts are a proxy for effort and a poor one, since they ignore expert time and the $38$ lines we write include a feasibility predicate that requires understanding the problem's constraints exactly and is the most safety-critical handwritten component.
They also exclude the roughly $30$-line repair pass of Section \ref{subsec:lastmile}.
Lastly, the comparison is per problem, so a user who solves one problem forever amortizes the scaffold and for them the argument in this paper is weak.

\section{What Gets Transformed, and Where It Breaks}\label{sec:analysis}

The results above treat the transformation step as a black box, and hence we dissect it in this section, i.e., what the model proposes, whether the score that selects among proposals means anything, and what happens as constraints tighten.

\subsection{Which Transformations Are Selected}\label{subsec:chosen}

We draw three observations from the transformations the framework retained, which Table~\ref{tab:chosen} lists per problem.
First, every retained target problem corresponds to a construction or decomposition an operations researcher would recognize, such as cluster-first-route-second for CVRP, the giant-tour split, and cutting stock for bin packing, and none is novel as an idea.
We take this as the expected result rather than a disappointment, since the model is asked to retrieve a representation and not to invent one, and the novelty in our results comes from the heuristics evolved inside those representations and from the transfer between them.
It also bounds the claim, which Section \ref{subsec:retrieval} tests.
Second, the fraction of candidates that survive the feasibility check at initialization runs from $90\%$ on KP down to $37\%$ on the last-mile task, in the same order as the number and kind of constraints each problem carries.
This single number is a useful early warning, since it is computable after roughly $19\%$ of the run.
Lastly, the score $s_j$ that selects among candidates correlates with the eventual test performance of the heuristics built on it at $\rho=0.69$ on pooled within-problem ranks ($p<0.001$, $n=105$ transformations; $0.58$--$0.78$ per problem), which is strong enough to justify selecting on it and far too weak to justify selecting only the best candidate, and that is one reason to retain $M=3$ candidates.

\begin{table*}[h]
    \centering
    \footnotesize
    \caption{The transformations retained across five runs per problem, paraphrased from the model's own descriptions.
    ``Runs'' is how many of the five runs retained a transformation of that family after selection, and since a run retains $M=3$, the counts in a block sum to $15$.
    ``Valid at init.'' is the fraction of the $M_0=10$ candidates in that problem whose transformation produced at least $\ell$ heuristics feasible on every training instance, averaged over runs.
    ``$\rho$'' is the Spearman rank correlation between a retained transformation's score $s_j$ at the end of initialization and the test objective of the best heuristic it eventually produced, over all retained transformations for that problem.}
    \begin{tabular}{l|l|c|c|c}
        \thickhline
        Problem & Target problems retained (paraphrased) & Runs & Valid at init.\ & $\rho$ \\\hline
        \multirow{3}{*}{TSP} & multi-start nearest-neighbor construction with candidate pruning & 5 & \multirow{3}{*}{87\%} & \multirow{3}{*}{0.78} \\
         & cheapest-insertion ordering into a growing partial tour & 5 & & \\
         & spanning-structure construction followed by shortcutting & 4 & & \\\hline
        \multirow{3}{*}{CVRP} & capacitated clustering, then ordering within each cluster & 5 & \multirow{3}{*}{63\%} & \multirow{3}{*}{0.71} \\
         & one giant tour split into capacity-feasible segments & 4 & & \\
         & assignment to a fixed vehicle count with load balancing, then ordering & 4 & & \\\hline
        \multirow{3}{*}{KP} & continuous relaxation by value density, with a repair step & 5 & \multirow{3}{*}{90\%} & \multirow{3}{*}{0.64} \\
         & greedy subset selection under one budget with pairwise exchange & 5 & & \\
         & layered shortest path over remaining capacity & 3 & & \\\hline
        \multirow{3}{*}{MKP} & item selection by aggregated value-to-weight ratio & 5 & \multirow{3}{*}{83\%} & \multirow{3}{*}{0.69} \\
         & generalized assignment of items to knapsacks & 5 & & \\
         & surrogate single-constraint knapsack by constraint aggregation & 4 & & \\\hline
        \multirow{3}{*}{BPP} & scheduling on identical machines under a load bound & 5 & \multirow{3}{*}{77\%} & \multirow{3}{*}{0.74} \\
         & cutting stock with pattern generation & 5 & & \\
         & interval selection on a residual-capacity structure & 3 & & \\\hline
        \multirow{3}{*}{OBPP} & online item-to-bin assignment scored by residual capacity & 5 & \multirow{3}{*}{70\%} & \multirow{3}{*}{0.58} \\
         & online assignment to machines with a fit-quality score & 5 & & \\
         & online matching between arriving items and open bins & 4 & & \\\hline
        \multirow{3}{*}{Last-mile} & routing with capacity only, time windows folded into travel cost & 5 & \multirow{3}{*}{37\%} & \multirow{3}{*}{0.66} \\
         & job scheduling with release dates, deadlines and setup times & 4 & & \\
         & time-expanded shortest path with a capacity resource & 3 & & \\
        \thickhline
    \end{tabular}
    \label{tab:chosen}
\end{table*}

\subsection{The Constraint-Tightness Ladder}\label{subsec:ladder}

The clearest way to find the boundary of the approach is to walk to it.
Starting from CVRP, we add one constraint family at a time, namely soft time windows enforced by a penalty, then hard time windows, then pickup-and-delivery precedence between paired stops, keeping every other setting fixed so that the only thing that varies is the shape of the feasible set (see Appendix~C.4).
As shown in Figure~\ref{fig:tightness}, feasibility is essentially free under a capacity constraint alone ($99.4\%$) and survives soft time windows ($96.1\%$), which is unsurprising, since a penalty term makes every solution feasible by definition and only moves the difficulty into the objective.
Hard time windows cut it to $57.4\%$ and adding precedence cuts it to $23.1\%$, and a stronger designer LLM shifts the whole curve up ($86.2\%$ and $41.8\%$) without changing its shape, so the problem is not solely a matter of model capability.
The objective gap moves in the opposite direction over a shrinking feasible subset, from $5.2\%$ to $21.7\%$, so the two curves compound rather than trade off.
That is, at the tight end the method returns worse solutions on fewer instances.

\begin{figure}[t]
    \centering
    \includegraphics[width=0.94\linewidth]{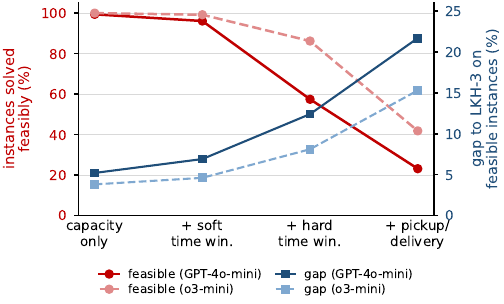}
    \caption{Adding one constraint at a time to CVRP.
    The red curves are the fraction of held-out instances on which the best synthesized transformation returns a feasible solution, and the blue curves are the objective gap to LKH-3 over feasible instances only.} 
    \label{fig:tightness}
\end{figure}

The same phenomenon appears along the search trajectory rather than across problems.
Violations do not decay with search budget on routing with hard time windows, but plateau above $40\%$ after a few hundred fitness evaluations and stay there, and relaxing the earliest-arrival bound, which lets vehicles serve customers early, moves the plateau to roughly $25\%$ without removing it (Figure~S4).

\subsection{How Transformations Fail}\label{subsec:taxonomy}

To investigate what went wrong, we sampled $50$ infeasible instance-transformation pairs at each of the two tight rungs and classified each by hand into one of three mechanisms, whose counts Figure~S5 reports.
In mechanism (a), $f$ builds an instance of the target problem that does not carry the constraint, i.e., time windows are omitted from the emitted structure, so no solution of $B$ can respect them.
This is the dominant mechanism under hard time windows ($27/50$) and is also the most tractable, since it is visible by inspecting $f$'s output alone and refinement occasionally repairs it.
In mechanism (b), $g$ produces an object of the right shape that nonetheless violates a constraint of $A$ ($16/50$ and $14/50$).
These are the dangerous ones, because nothing short of the feasibility predicate detects them, which is the concrete reason Section \ref{subsec:residual} keeps that predicate handcrafted.
In mechanism (c), the model selects a target problem whose feasible set has no room for the constraint at all, for instance a pure sequencing formulation for a problem with precedence between paired stops ($7/50$ and $17/50$), and no better implementation of $f$ and $g$ helps when the failure is in the choice of $B$.

The shift in the mixture between the two rungs is the finding.
Under hard time windows most failures are implementation errors, which better models and refinement should keep eroding, whereas once precedence is added, expressiveness failures become the largest single category, which scaling the LLM will not fix.

\subsection{Retrieval Versus Design}\label{subsec:retrieval}

Section \ref{subsec:chosen} shows the model retrieving well-known formulations, which raises the question of what happens on a problem it has not seen described in familiar terms.
For each of five problems we therefore wrote a second prose description that is mathematically equivalent to the first but uses no standard terminology, no problem name, and a re-skinned cover story, and reran the framework unchanged.
From Table~S12, we observe that performance degrades, and the degradation is informative in its distribution.
It is nil or slightly positive on the two knapsack-family problems and small on TSP ($1.4\%$), whereas on CVRP it is $5.9\%$ and on the last-mile task feasibility falls from $61.3\%$ to $38.5\%$, with the fraction of candidates valid at initialization falling from $78\%$ to $51\%$ over the five problems.
The pattern is that obfuscation costs little when the target problem is reachable from the structure of the description itself, and a great deal when the useful transformation depends on recognizing a named formulation.
We report this as a real limitation rather than as a robustness result, since a substantial part of what the transformation step contributes is retrieval, and on genuinely unfamiliar problems it retrieves less.

\section{Limitations}\label{sec:limit}

First, validity is empirically tested and not proven, as Definition \ref{def:transformation} quantifies over all instances whereas our check covers $n_{\mathrm{tr}}$ of them, and Section \ref{subsec:ladder} shows that violations on unseen instances are common rather than rare once constraints tighten.
Any deployment must therefore keep an independent feasibility check on the output. 

Second, our approach is bounded by what the model represents well, as Sections \ref{subsec:taxonomy} and \ref{subsec:retrieval} show.
On a genuinely new problem, one under-studied in the literature the model was trained on, we expect the transformation step to degenerate to the identity, in which case the framework reduces to running an evolutionary search directly on $A$ with no scaffold, which is worse than running it with one.
A practical mitigation, not evaluated here, is to spend a stronger model on initialization only.
Initialization is $19\%$ of wall-clock and under $10\%$ of LLM calls (Table~S8), so the cost of doing so is small.

Third, manual efforts are reduced but not eliminated, since the prose description and the feasibility predicate remain handcrafted (Section \ref{subsec:residual}) and on the last-mile task we had to add a repair pass to reach full feasibility (Section \ref{subsec:lastmile}).

Lastly, we use only five runs, and most baselines are reproduced from the literature and cover only the two skeletons used there, not a portfolio of representations or another automatic outer loop.

\section{Conclusion}\label{sec:conclu}

Evolutionary code search frameworks have made the search over programs much stronger while leaving untouched the manual work of deciding what the search may edit.
In this paper, we seek to automate that decision.
By having an LLM propose a transformation of the problem at hand into related problems it represents well, the specification an evolutionary search needs can be synthesized rather than written, and heuristics can then be evolved for several such target problems jointly so that ideas migrate between representations.
On seven problems this is mostly competitive with or better than heuristics designed inside handcrafted frameworks while cutting handcrafted scaffold lines by roughly a factor of six, and the synthesized specification drives four different searches unchanged, though the most constrained problem still needs handcrafted repair.
Future studies may delve deeper into feasibility preservation techniques, e.g., property-based testing or bounded verification of the synthesized maps, and run user studies of the effort saved.

\begin{acks}
Our project is supported by the Rural Drug Addiction Research Center at the University of Nebraska-Lincoln, and the National Science Foundation under grant IIS:RI \#2302999 and IIS:RI \#2414554.
The content is solely the responsibility of the authors and does not necessarily represent the official views of the funding agencies.
\end{acks}

\section*{Ethical Considerations}\label{sec:ethics}

\paragraph{Correctness and Safe Deployment.}
The central risk in our work is that a transformation which passed every check we apply can still return infeasible solutions on instances it has not seen.
We measure how often this happens.
Section \ref{subsec:ladder} reports feasibility falling to $23\%$ of instances on the tightest setting we tested, and Section \ref{subsec:taxonomy} identifies a failure mode in which the returned object looks structurally correct and violates a constraint silently.
In an operational setting such as vehicle routing under delivery commitments, an infeasible route that is not detected as such is worse than no route at all.
We therefore recommend that any deployment keep an independent feasibility check between the designed heuristic and the world, written and maintained separately from the framework, and treat a drop in the empirical validity rate as an operational alarm rather than as noise.
Our work should not be read as evidence that LLM-generated transformations can be relied on without such a check.

\paragraph{Compute and Monetary Cost.}
Designing a heuristic with our framework costs under $\$0.15$ and under two hours on a single workstation per problem with a small model, and roughly $6$--$7\times$ that with a reasoning model (Table~S8).
This is modest, and we report it in full so that it can be accurately compared.
The resulting heuristics run one to three orders of magnitude faster per instance than the solvers they are compared against (Table \ref{tab:solver-ref}), so in settings where the same problem is solved repeatedly, the design cost is recovered quickly in inference energy.
We would caution against reading that as an unambiguous environmental benefit.
Making heuristic design cheap also makes it easier to run many searches, and aggregate consumption depends on how the capability is used.


\paragraph{Data, Human Subjects, and Dual Use.}
The experiments use synthetic instance generators, the public TSPLib and Gehring--Homberger benchmark sets, and the 2021 Amazon Last Mile Routing Research Challenge data set \citep{merchan2024amazon}, which is published for research use with route and package identifiers regenerated and location data obfuscated by its providers.
We use it under those terms and derive no individual-level information from it.
No human subjects were involved and no personal data was collected or processed.
The optimization problems studied here are general-purpose logistics and packing problems, and we are not aware of a dual-use concern specific to this work beyond those that apply to optimization tooling in general.

\bibliographystyle{ACM-Reference-Format}
\bibliography{main}

@inproceedings{liu2024evolution,
  title={Evolution of Heuristics: Towards Efficient Automatic Algorithm Design Using Large Language Model},
  author={Liu, Fei and Xialiang, Tong and Yuan, Mingxuan and Lin, Xi and Luo, Fu and Wang, Zhenkun and Lu, Zhichao and Zhang, Qingfu},
  booktitle={Forty-first International Conference on Machine Learning},
  year={2024}
}

@inproceedings{zheng2025monte,
  title={Monte Carlo Tree Search for Comprehensive Exploration in LLM-Based Automatic Heuristic Design},
  author={Zheng, Zhi and Xie, Zhuoliang and Wang, Zhenkun and Hooi, Bryan},
  booktitle={Forty-Second International Conference on Machine Learning},
  year={2025}
}

@article{romera2024mathematical,
  title={Mathematical discoveries from program search with large language models},
  author={Romera-Paredes, Bernardino and Barekatain, Mohammadamin and Novikov, Alexander and Balog, Matej and Kumar, M Pawan and Dupont, Emilien and Ruiz, Francisco JR and Ellenberg, Jordan S and Wang, Pengming and Fawzi, Omar and others},
  journal={Nature},
  volume={625},
  number={7995},
  pages={468--475},
  year={2024},
  publisher={Nature Publishing Group UK London}
}

@inproceedings{ye2024reevo,
    title={ReEvo: Large Language Models as Hyper-Heuristics with Reflective Evolution}, 
    author={Ye, Haoran and Wang, Jiarui and Cao, Zhiguang and Berto, Federico and Hua, Chuanbo and Kim, Haeyeon and Park, Jinkyoo and Song, Guojie},
    booktitle={Advances in Neural Information Processing Systems},
    year={2024},
    note={\url{https://github.com/ai4co/reevo}}
}

@misc{yang2025nondeterministic,
      title={Nondeterministic Polynomial-time Problem Challenge: An Ever-Scaling Reasoning Benchmark for LLMs}, 
      author={Chang Yang and Ruiyu Wang and Junzhe Jiang and Qi Jiang and Qinggang Zhang and Yanchen Deng and Shuxin Li and Shuyue Hu and Bo Li and Florian T. Pokorny and Xiao Huang and Xinrun Wang},
      year={2025},
      eprint={2504.11239},
      archivePrefix={arXiv},
      primaryClass={cs.AI},
      url={https://arxiv.org/abs/2504.11239}, 
}

@inproceedings{yao2024multi,
  title={Multi-objective evolution of heuristic using large language model},
  author={Yao, Shunyu and Liu, Fei and Lin, Xi and Lu, Zhichao and Wang, Zhenkun and Zhang, Qingfu},
  booktitle={Proceedings of the AAAI Conference on Artificial Intelligence},
  volume={39},
  number={25},
  pages={27144--27152},
  year={2025}
}

@inproceedings{liu2024large,
  title={Large language models as evolutionary optimizers},
  author={Liu, Shengcai and Chen, Caishun and Qu, Xinghua and Tang, Ke and Ong, Yew-Soon},
  booktitle={2024 IEEE Congress on Evolutionary Computation (CEC)},
  pages={1--8},
  year={2024},
  organization={IEEE}
}

@inproceedings{crescenzi1997short,
  title={A short guide to approximation preserving reductions},
  author={Crescenzi, Pierluigi},
  booktitle={Proceedings of Computational Complexity. Twelfth Annual IEEE Conference},
  pages={262--273},
  year={1997},
  organization={IEEE}
}

@book{pillay2018hyper,
  title={Hyper-heuristics: theory and applications},
  author={Pillay, Nelishia and Qu, Rong},
  year={2018},
  publisher={Springer}
}

@article{drake2020recent,
  title={Recent advances in selection hyper-heuristics},
  author={Drake, John H and Kheiri, Ahmed and {\"O}zcan, Ender and Burke, Edmund K},
  journal={European Journal of Operational Research},
  volume={285},
  number={2},
  pages={405--428},
  year={2020},
  publisher={Elsevier}
}

@inproceedings{duflo2019gp,
  title={A gp hyper-heuristic approach for generating tsp heuristics},
  author={Duflo, Gabriel and Kieffer, Emmanuel and Brust, Matthias R and Danoy, Gr{\'e}goire and Bouvry, Pascal},
  booktitle={2019 IEEE International Parallel and Distributed Processing Symposium Workshops (IPDPSW)},
  pages={521--529},
  year={2019},
  organization={IEEE}
}

@article{zhao2023automated,
  title={Automated design of metaheuristic algorithms: A survey},
  author={Zhao, Qi and Duan, Qiqi and Yan, Bai and Cheng, Shi and Shi, Yuhui},
  journal={arXiv preprint arXiv:2303.06532},
  year={2023}
}

@article{reinelt1991tsplib,
  title={TSPLIB—A traveling salesman problem library},
  author={Reinelt, Gerhard},
  journal={ORSA journal on computing},
  volume={3},
  number={4},
  pages={376--384},
  year={1991},
  publisher={Informs}
}

@article{burke2013hyper,
  title={Hyper-heuristics: A survey of the state of the art},
  author={Burke, Edmund K and Gendreau, Michel and Hyde, Matthew and Kendall, Graham and Ochoa, Gabriela and {\"O}zcan, Ender and Qu, Rong},
  journal={Journal of the Operational Research Society},
  volume={64},
  number={12},
  pages={1695--1724},
  year={2013},
  publisher={Taylor \& Francis}
}

@book{langdon2013foundations,
  title={Foundations of genetic programming},
  author={Langdon, William B and Poli, Riccardo},
  year={2013},
  publisher={Springer Science \& Business Media}
}

@article{mahowald2024dissociating,
  title={Dissociating language and thought in large language models},
  author={Mahowald, Kyle and Ivanova, Anna A and Blank, Idan A and Kanwisher, Nancy and Tenenbaum, Joshua B and Fedorenko, Evelina},
  journal={Trends in cognitive sciences},
  year={2024},
  publisher={Elsevier}
}

@article{dorigo2007ant,
  title={Ant colony optimization},
  author={Dorigo, Marco and Birattari, Mauro and Stutzle, Thomas},
  journal={IEEE computational intelligence magazine},
  volume={1},
  number={4},
  pages={28--39},
  year={2007},
  publisher={IEEE}
}

@incollection{voudouris2010guided,
  title={Guided local search},
  author={Voudouris, Christos and Tsang, Edward PK and Alsheddy, Abdullah},
  booktitle={Handbook of metaheuristics},
  pages={321--361},
  year={2010},
  publisher={Springer}
}

@article{o2010open,
  title={Open issues in genetic programming},
  author={O’Neill, Michael and Vanneschi, Leonardo and Gustafson, Steven and Banzhaf, Wolfgang},
  journal={Genetic Programming and Evolvable Machines},
  volume={11},
  pages={339--363},
  year={2010},
  publisher={Springer}
}

@article{bengio2021machine,
  title={Machine learning for combinatorial optimization: a methodological tour d’horizon},
  author={Bengio, Yoshua and Lodi, Andrea and Prouvost, Antoine},
  journal={European Journal of Operational Research},
  volume={290},
  number={2},
  pages={405--421},
  year={2021},
  publisher={Elsevier}
}

@inproceedings{yang2023survey,
  title={A survey on reinforcement learning for combinatorial optimization},
  author={Yang, Yunhao and Whinston, Andrew},
  booktitle={2023 IEEE World Conference on Applied Intelligence and Computing (AIC)},
  pages={131--136},
  year={2023},
  organization={IEEE}
}

@article{zheng2024udc,
  title={UDC: A unified neural divide-and-conquer framework for large-scale combinatorial optimization problems},
  author={Zheng, Zhi and Zhou, Changliang and Xialiang, Tong and Yuan, Mingxuan and Wang, Zhenkun},
  journal={arXiv preprint arXiv:2407.00312},
  year={2024}
}

@article{kwon2020pomo,
  title={Pomo: Policy optimization with multiple optima for reinforcement learning},
  author={Kwon, Yeong-Dae and Choo, Jinho and Kim, Byoungjip and Yoon, Iljoo and Gwon, Youngjune and Min, Seungjai},
  journal={Advances in Neural Information Processing Systems},
  volume={33},
  pages={21188--21198},
  year={2020}
}

@article{ye2023deepaco,
  title={DeepACO: Neural-enhanced ant systems for combinatorial optimization},
  author={Ye, Haoran and Wang, Jiarui and Cao, Zhiguang and Liang, Helan and Li, Yong},
  journal={Advances in neural information processing systems},
  volume={36},
  pages={43706--43728},
  year={2023}
}

@software{ortools_routing,
  title = {OR-Tools Routing Library},
  version = { v9.11 },
  author = {Vincent Furnon and Laurent Perron},
  organization = {Google},
  url = {https://developers.google.com/optimization/routing/},
  date = { 2024-05-07 }
}

@article{mei2022explainable,
  title={Explainable artificial intelligence by genetic programming: A survey},
  author={Mei, Yi and Chen, Qi and Lensen, Andrew and Xue, Bing and Zhang, Mengjie},
  journal={IEEE Transactions on Evolutionary Computation},
  volume={27},
  number={3},
  pages={621--641},
  year={2022},
  publisher={IEEE}
}

@article{huang2025survey,
  title={A survey on hallucination in large language models: Principles, taxonomy, challenges, and open questions},
  author={Huang, Lei and Yu, Weijiang and Ma, Weitao and Zhong, Weihong and Feng, Zhangyin and Wang, Haotian and Chen, Qianglong and Peng, Weihua and Feng, Xiaocheng and Qin, Bing and others},
  journal={ACM Transactions on Information Systems},
  volume={43},
  number={2},
  pages={1--55},
  year={2025},
  publisher={ACM New York, NY}
}

@article{rosenkrantz1977analysis,
  title={An analysis of several heuristics for the traveling salesman problem},
  author={Rosenkrantz, Daniel J and Stearns, Richard E and Lewis, II, Philip M},
  journal={SIAM journal on computing},
  volume={6},
  number={3},
  pages={563--581},
  year={1977},
  publisher={SIAM}
}

@article{brecklinghaus2015approximation,
  title={The approximation ratio of the greedy algorithm for the metric traveling salesman problem},
  author={Brecklinghaus, Judith and Hougardy, Stefan},
  journal={Operations Research Letters},
  volume={43},
  number={3},
  pages={259--261},
  year={2015},
  publisher={Elsevier}
}

@inproceedings{asani2023computation,
  title={A computation investigation of the impact of convex hull subtour on the nearest neighbour heuristic},
  author={Asani, Emmanuel O and Okeyinka, Aderemi E and Adebiyi, Ayodele Ariyo},
  booktitle={2023 International Conference on Science, Engineering and Business for Sustainable Development Goals (SEB-SDG)},
  volume={1},
  pages={1--7},
  year={2023},
  organization={IEEE}
}

@article{liu2024llm4ad,
    title = {LLM4AD: A Platform for Algorithm Design with Large Language Model},
    author = {Fei Liu and Rui Zhang and Zhuoliang Xie and Rui Sun and Kai Li and Xi Lin and Zhenkun Wang and Zhichao Lu and Qingfu Zhang},
    year = {2024},
    eprint = {2412.17287},
    archivePrefix = {arXiv},
    primaryClass = {cs.AI},
    url = {https://arxiv.org/abs/2412.17287},
}

@article{novikov2025alphaevolve,
  title={Alphaevolve: A coding agent for scientific and algorithmic discovery},
  author={Novikov, Alexander and V{\~u}, Ng{\^a}n and Eisenberger, Marvin and Dupont, Emilien and Huang, Po-Sen and Wagner, Adam Zsolt and Shirobokov, Sergey and Kozlovskii, Borislav and Ruiz, Francisco JR and Mehrabian, Abbas and others},
  journal={arXiv preprint arXiv:2506.13131},
  year={2025}
}

@software{sharma2025openevolve,
  title = {OpenEvolve: an open-source evolutionary coding agent},
  author = {Asankhaya Sharma},
  year = {2025},
  publisher = {GitHub},
  url = {https://github.com/algorithmicsuperintelligence/openevolve}
}

@inproceedings{lange2026shinkaevolve,
  title={Shinkaevolve: Towards open-ended and sample-efficient program evolution},
  author={Lange, Robert and Imajuku, Yuki and Cetin, Edoardo},
  booktitle={International Conference on Learning Representations},
  volume={2026},
  pages={74026--74078},
  year={2026}
}

@article{assumpccao2025codeevolve,
  title={CodeEvolve: an open source evolutionary coding agent for algorithmic discovery and optimization},
  author={Assump{\c{c}}{\~a}o, Henrique and Ferreira, Diego and Campos, Leandro and Murai, Fabricio},
  journal={arXiv preprint arXiv:2510.14150},
  year={2025}
}

@techreport{christofides1976worst,
  title={Worst-case analysis of a new heuristic for the travelling salesman problem},
  author={Christofides, Nicos},
  year={1976},
  number={388},
  institution={Graduate School of Industrial Administration, Carnegie Mellon University},
}

@article{helsgaun2017extension,
  title={An extension of the Lin-Kernighan-Helsgaun TSP solver for constrained traveling salesman and vehicle routing problems},
  author={Helsgaun, Keld},
  journal={Roskilde: Roskilde University},
  volume={12},
  number={966-980},
  pages={1},
  year={2017}
}

@book{applegate2006traveling,
  title={The Traveling Salesman Problem: A Computational Study},
  author={Applegate, D.L.},
  isbn={9780691129938},
  lccn={2006931528},
  series={Princeton Series in Applied Mathematics},
  url={https://books.google.com/books?id=vhsJbqomDuIC},
  year={2006},
  publisher={Princeton University Press}
}

@article{solomon1987algorithms,
  title={Algorithms for the vehicle routing and scheduling problems with time window constraints},
  author={Solomon, Marius M},
  journal={Operations research},
  volume={35},
  number={2},
  pages={254--265},
  year={1987},
  publisher={Informs}
}

@inproceedings{gehring1999parallel,
  title={A parallel hybrid evolutionary metaheuristic for the vehicle routing problem with time windows},
  author={Gehring, Hermann and Homberger, J{\"o}rg},
  booktitle={Proceedings of EUROGEN99},
  volume={2},
  pages={57--64},
  year={1999},
  organization={Springer Berlin}
}

@article{merchan2024amazon,
  title={2021 Amazon last mile routing research challenge: Data set},
  author={Merch{\'a}n, Daniel and Arora, Jatin and Pachon, Julian and Konduri, Karthik and Winkenbach, Matthias and Parks, Steven and Noszek, Joseph},
  journal={Transportation Science},
  volume={58},
  number={1},
  pages={8--11},
  year={2024},
  publisher={INFORMS}
}

@article{vidal2022hybrid,
  title={Hybrid genetic search for the CVRP: Open-source implementation and SWAP* neighborhood},
  author={Vidal, Thibaut},
  journal={Computers \& Operations Research},
  volume={140},
  pages={105643},
  year={2022},
  publisher={Elsevier}
}

@article{liu2026systematic,
author = {Liu, Fei and Yao, Yiming and Guo, Ping and Yang, Zhiyuan and Lin, Xi and Zhao, Zhe and Tong, Xialiang and Mao, Kun and Lu, Zhichao and Wang, Zhenkun and Yuan, Mingxuan and Zhang, Qingfu},
title = {A Systematic Survey on Large Language Models for Algorithm Design},
year = {2026},
issue_date = {June 2026},
publisher = {Association for Computing Machinery},
address = {New York, NY, USA},
volume = {58},
number = {8},
issn = {0360-0300},
url = {https://doi.org/10.1145/3787585},
doi = {10.1145/3787585},
journal = {ACM Comput. Surv.},
month = feb,
articleno = {218},
numpages = {32},
}

\clearpage
\onecolumn

\appendix

\setcounter{figure}{0}
\setcounter{table}{0}
\setcounter{equation}{0}
\renewcommand{\thefigure}{S\arabic{figure}}
\renewcommand{\thetable}{S\arabic{table}}
\renewcommand{\theequation}{S\arabic{equation}}

\section{Considered COPs}\label{app:cops}

In this appendix, we introduce the considered COPs and define the objective function for each ($q$ in Equations \ref{eq:prob-def} and \ref{eq:prob-def-real}).
We follow the problem definitions and setups from \citet{zheng2025monte} (which followed \citet{ye2024reevo}).
TSP, CVRP, BPP, and OBPP are minimization problems while KP and MKP are maximization problems.

\paragraph{Traveling Salesman Problem (TSP).}
TSP aims to find the shortest path to visit each of the $n$ nodes once and return to the starting node. Each TSP instance contains the Euclidean distance matrix $\bm{D}$ where $d_{ij}$ denotes the cost between node $i$ and $j$.
The solution of TSP is a permutation of all node indices $\bm{s}=(s_1,s_2,\ldots,s_n)$.
Thus, the (negated) objective function is
\begin{equation*}
    -\left(\sum_{t=1}^{n-1}d_{s_t,s_{t+1}}+d_{s_n,s_1}\right).
\end{equation*}

\paragraph{Capacitated Vehicle Routing Problem (CVRP).}
CVRP aims to plan several capacity-constrained vehicles starting at and returning to a depot, meeting the demands of multiple customers, and minimizing the total travel distance. Each CVRP instance contains a depot (the 0-th node) and $n$ customers.
Let $\bm{D}$ be the Euclidean distance matrix. The (negated) objective function is
\begin{equation*}
\begin{array}{ll}
& -\sum_{j=1}^{q} C\left(\boldsymbol{\varrho}^{j}\right), \\
& C\left(\boldsymbol{\varrho}^{j}\right)=\sum_{t=0}^{\left|\boldsymbol{\varrho}^{j}\right|-1} d_{\varrho_{t}^{j}, \varrho_{t+1}^{j}}^{j}+d_{\varrho_{n_{j}}^{j}, \varrho_{0}^{j}}, \\
\text { s.t. } & 0 \leq \delta_{i} \leq C, \quad \sum_{i \in \boldsymbol{\varrho}^{j}} \delta_{i} \leq C, \quad i \in\{1, \ldots, n\}, j \in\{1, \ldots, q\},
\end{array}
\end{equation*}
where $\bm{s}$ is a solution representing the complete route of vehicles and consists of $q$ sub-routes $\bm{s} = \{\bm{\varrho}^1, \bm{\varrho}^2,\ldots,\bm{\varrho}^q\}$. Each sub-route $\bm{\varrho}^j = (\bm{\varrho}^j_1,\ldots, \bm{\varrho}^j_{n_j} )$, $j\in \{1,\ldots,q\}$ starts from the depot $s_0$ and goes back to $s_0$; $n_j$ represents the number of customer nodes in it such that $n = \sum^q_{j=1} n_j$. $\delta_i$ denotes the demand of node $i$, and $C$ denotes the capacity of the vehicles.

\paragraph{(Offline) Bin Packing Problem (BPP).}
BPP aims to place a set of $n$ items with different sizes into as few bins as possible, each of which has capacity of $W$.
The solution of BPP is $\bm{s} = \{\bm{s}^1, \bm{s}^2,\ldots,\bm{s}^K\}$ where $\bm{s}^i$ is the set of item indices for the $i$-th bin and $K$ is the number of bins used. The (negated) objective function is
\begin{equation*}
\begin{array}{ll}
& -K, \\
\text { s.t. } & \sum_{j\in\bm{s}^i} w_j \le W, \quad i \in\{1, \ldots, K\}.
\end{array}
\end{equation*}

\paragraph{Online Bin Packing Problem (OBPP).}
OBPP additionally requires making an immediate decision on which bin to place once a new item arrives, without any information on future items. The objective function is similar to BPP.

\paragraph{0/1 Knapsack Problem (KP).}
KP aims to pack items of maximum total value to a knapsack with capacity $W$. Each of the $n$ available items can only be picked once. The solution of KP is the set of indices of the selected items $\bm{s}\subseteq\{1,2,\ldots,n\}$. Let $w_j$ and $v_j$ be the weight and value of item $j$, respectively. The objective function is 
\begin{equation*}
\begin{array}{ll}
& \sum_{j\in\bm{s}} v_j, \\
\text { s.t. } & \sum_{j\in\bm{s}} w_j \le W.
\end{array}
\end{equation*}

\paragraph{Multiple Knapsack Problem (MKP).}
MKP extends KP to $m>1$ knapsacks. The solution of MKP is now $\bm{s} = \{\bm{s}^1, \bm{s}^2,\ldots,\bm{s}^m\}$ where $\bm{s}^i$ is the set of indices of the selected items for the $i$-th knapsack. The objective function is
\begin{equation*}
\begin{array}{ll}
& \sum_{i=1}^m\sum_{j\in\bm{s}^i} v_{j}, \\
\text { s.t. } & \sum_{j\in\bm{s}^i} w_j \le W_i, \quad i \in\{1, \ldots, m\}.
\end{array}
\end{equation*}

\newpage
\section{How ACO Is Employed in Prior Evolutionary Code Search Works}\label{app:aco}

As described in \citet{zheng2025monte}, ACO is an evolutionary algorithm inspired by the behavior of ants to find the shortest route between their colony and food sources \citep{dorigo2007ant}.

ACO records a pheromone matrix $\bm{\tau}$ and a heuristic matrix $\bm{\eta}$. Each item $\tau_{ij}$ in $\bm{\tau}$ indicates the priority of including an edge $(i, j)$ in a solution. The pheromone trails are iteratively updated based on the quality of the solutions found, encouraging future ants to follow better paths. The heuristic information on each edge, i.e., $\eta_{ij}$, is a problem-specific measure that indicates the immediate benefit of choosing a particular path. For solving TSP with ACO, for example, $\eta_{ij}$ is often set to be the inverse of the distance between cities $i$ and $j$, i.e., $\eta_{ij} = 1/d_{ij}$. In response, evolutionary code searches aim to design a more effective heuristic matrix $\bm{\eta}$ based on the problem-specific inputs.

Given $\bm{\eta}$, the virtual ants then construct solutions by moving from node to node, probabilistically choosing the next node based on a combination of pheromone and heuristic information. After all the ants have constructed their solutions, the pheromone levels update. An ACO iteration typically involves solution construction, optional local search, and pheromone update. By iteratively applying these steps, ACO algorithms can effectively explore the solution space and converge toward optimal or near-optimal solutions for COPs.

\paragraph{Implementation.}
The following listings respectively show the Python implementation of ACO for TSP and CVRP in both ReEvo \citep{ye2024reevo} and MCTS-AHD \citep{zheng2025monte}.
Albeit using the same algorithmic framework, there are substantial differences between the two pieces of code, which means significant manual efforts are necessary when adopting ACO (and other algorithmic frameworks in general) for a particular COP.

\begin{lstlisting}[basicstyle=\tiny\ttfamily, caption={Implementation of the ACO framework for TSP in ReEvo \citep{ye2024reevo} and MCTS-AHD \citep{zheng2025monte}.},captionpos=b]
import torch
from torch.distributions import Categorical

class ACO():

    def __init__(self, 
                 distances,
                 heuristic,
                 n_ants=30, 
                 decay=0.9,
                 alpha=1,
                 beta=1,
                 device='cpu'
                 ):
        
        self.problem_size = len(distances)
        self.distances  = torch.tensor(distances, device=device) if not isinstance(distances, torch.Tensor) else distances
        self.n_ants = n_ants
        self.decay = decay
        self.alpha = alpha
        self.beta = beta
        
        self.pheromone = torch.ones_like(self.distances)
        self.heuristic = torch.tensor(heuristic, device=device) if not isinstance(heuristic, torch.Tensor) else heuristic

        self.shortest_path = None
        self.lowest_cost = float('inf')

        self.device = device

    @torch.no_grad()
    def run(self, n_iterations):
        for _ in range(n_iterations):
            paths = self.gen_path(require_prob=False)
            costs = self.gen_path_costs(paths)
            
            best_cost, best_idx = costs.min(dim=0)
            if best_cost < self.lowest_cost:
                self.shortest_path = paths[:, best_idx]
                self.lowest_cost = best_cost
            
            self.update_pheronome(paths, costs)

        return self.lowest_cost
       
    @torch.no_grad()
    def update_pheronome(self, paths, costs):
        '''
        Args:
            paths: torch tensor with shape (problem_size, n_ants)
            costs: torch tensor with shape (n_ants,)
        '''
        self.pheromone = self.pheromone * self.decay 
        for i in range(self.n_ants):
            path = paths[:, i]
            cost = costs[i]
            self.pheromone[path, torch.roll(path, shifts=1)] += 1.0/cost
            self.pheromone[torch.roll(path, shifts=1), path] += 1.0/cost

    @torch.no_grad()
    def gen_path_costs(self, paths):
        '''
        Args:
            paths: torch tensor with shape (problem_size, n_ants)
        Returns:
                Lengths of paths: torch tensor with shape (n_ants,)
        '''
        assert paths.shape == (self.problem_size, self.n_ants)
        u = paths.T # shape: (n_ants, problem_size)
        v = torch.roll(u, shifts=1, dims=1)  # shape: (n_ants, problem_size)
        assert (self.distances[u, v] > 0).all()
        return torch.sum(self.distances[u, v], dim=1)

    def gen_path(self, require_prob=False):
        '''
        Tour contruction for all ants
        Returns:
            paths: torch tensor with shape (problem_size, n_ants), paths[:, i] is the constructed tour of the ith ant
            log_probs: torch tensor with shape (problem_size, n_ants), log_probs[i, j] is the log_prob of the ith action of the jth ant
        '''
        start = torch.randint(low=0, high=self.problem_size, size=(self.n_ants,), device=self.device)
        mask = torch.ones(size=(self.n_ants, self.problem_size), device=self.device)
        mask[torch.arange(self.n_ants, device=self.device), start] = 0
        
        paths_list = [] # paths_list[i] is the ith move (tensor) for all ants
        paths_list.append(start)
        
        log_probs_list = [] # log_probs_list[i] is the ith log_prob (tensor) for all ants' actions
        
        prev = start
        for _ in range(self.problem_size-1):
            actions, log_probs = self.pick_move(prev, mask, require_prob)
            paths_list.append(actions)
            if require_prob:
                log_probs_list.append(log_probs)
                mask = mask.clone()
            prev = actions
            mask[torch.arange(self.n_ants, device=self.device), actions] = 0
        
        if require_prob:
            return torch.stack(paths_list), torch.stack(log_probs_list)
        else:
            return torch.stack(paths_list)
        
    def pick_move(self, prev, mask, require_prob):
        '''
        Args:
            prev: tensor with shape (n_ants,), previous nodes for all ants
            mask: bool tensor with shape (n_ants, p_size), masks (0) for the visited cities
        '''
        pheromone = self.pheromone[prev] # shape: (n_ants, p_size)
        heuristic = self.heuristic[prev] # shape: (n_ants, p_size)
        dist = ((pheromone ** self.alpha) * (heuristic ** self.beta) * mask) # shape: (n_ants, p_size)
        dist = Categorical(dist)
        actions = dist.sample() # shape: (n_ants,)
        log_probs = dist.log_prob(actions) if require_prob else None # shape: (n_ants,)
        return actions, log_probs
\end{lstlisting}

\begin{lstlisting}[basicstyle=\tiny\ttfamily, caption={Implementation of the ACO framework for CVRP in ReEvo \citep{ye2024reevo} and MCTS-AHD \citep{zheng2025monte}.},captionpos=b]
import torch
from torch.distributions import Categorical
import random
import itertools
import numpy as np

class ACO():
    def __init__(self,  # 0: depot
                 distances, # (n, n)
                 demand,   # (n, )
                 heuristic, # (n, n)
                 capacity,
                 n_ants=30, 
                 decay=0.9,
                 alpha=1,
                 beta=1,
                 device='cpu',
                 ):
        
        self.problem_size = len(distances)
        self.distances = torch.tensor(distances, device=device) if not isinstance(distances, torch.Tensor) else distances
        self.demand = torch.tensor(demand, device=device) if not isinstance(demand, torch.Tensor) else demand
        self.capacity = capacity
                
        self.n_ants = n_ants
        self.decay = decay
        self.alpha = alpha
        self.beta = beta
        
        self.pheromone = torch.ones_like(self.distances)
        self.heuristic = torch.tensor(heuristic, device=device) if not isinstance(heuristic, torch.Tensor) else heuristic

        self.shortest_path = None
        self.lowest_cost = float('inf')

        self.device = device
        

    @torch.no_grad()
    def run(self, n_iterations):
        for _ in range(n_iterations):
            paths = self.gen_path()
            costs = self.gen_path_costs(paths)
            
            best_cost, best_idx = costs.min(dim=0)
            if best_cost < self.lowest_cost:
                self.shortest_path = paths[:, best_idx]
                self.lowest_cost = best_cost
       
            self.update_pheronome(paths, costs)

        return self.lowest_cost
       
    @torch.no_grad()
    def update_pheronome(self, paths, costs):
        '''
        Args:
            paths: torch tensor with shape (problem_size, n_ants)
            costs: torch tensor with shape (n_ants,)
        '''
        self.pheromone = self.pheromone * self.decay 
        for i in range(self.n_ants):
            path = paths[:, i]
            cost = costs[i]
            self.pheromone[path[:-1], torch.roll(path, shifts=-1)[:-1]] += 1.0/cost
        self.pheromone[self.pheromone < 1e-10] = 1e-10
    
    @torch.no_grad()
    def gen_path_costs(self, paths):
        u = paths.permute(1, 0) # shape: (n_ants, max_seq_len)
        v = torch.roll(u, shifts=-1, dims=1)  
        return torch.sum(self.distances[u[:, :-1], v[:, :-1]], dim=1)

    def gen_path(self):
        actions = torch.zeros((self.n_ants,), dtype=torch.long, device=self.device)
        visit_mask = torch.ones(size=(self.n_ants, self.problem_size), device=self.device)
        visit_mask = self.update_visit_mask(visit_mask, actions)
        used_capacity = torch.zeros(size=(self.n_ants,), device=self.device)
        
        used_capacity, capacity_mask = self.update_capacity_mask(actions, used_capacity)
        
        paths_list = [actions] # paths_list[i] is the ith move (tensor) for all ants
        
        done = self.check_done(visit_mask, actions)
        while not done:
            actions = self.pick_move(actions, visit_mask, capacity_mask)
            paths_list.append(actions)
            visit_mask = self.update_visit_mask(visit_mask, actions)
            used_capacity, capacity_mask = self.update_capacity_mask(actions, used_capacity)
            done = self.check_done(visit_mask, actions)
            
        return torch.stack(paths_list)
        
    def pick_move(self, prev, visit_mask, capacity_mask):
        pheromone = self.pheromone[prev] # shape: (n_ants, p_size)
        heuristic = self.heuristic[prev] # shape: (n_ants, p_size)
        dist = ((pheromone ** self.alpha) * (heuristic ** self.beta) * visit_mask * capacity_mask) # shape: (n_ants, p_size)
        dist = Categorical(dist)
        actions = dist.sample() # shape: (n_ants,)
        return actions
    
    def update_visit_mask(self, visit_mask, actions):
        visit_mask[torch.arange(self.n_ants, device=self.device), actions] = 0
        visit_mask[:, 0] = 1 # depot can be revisited with one exception
        visit_mask[(actions==0) * (visit_mask[:, 1:]!=0).any(dim=1), 0] = 0 # one exception is here
        return visit_mask
    
    def update_capacity_mask(self, cur_nodes, used_capacity):
        '''
        Args:
            cur_nodes: shape (n_ants, )
            used_capacity: shape (n_ants, )
            capacity_mask: shape (n_ants, p_size)
        Returns:
            ant_capacity: updated capacity
            capacity_mask: updated mask
        '''
        capacity_mask = torch.ones(size=(self.n_ants, self.problem_size), device=self.device)
        # update capacity
        used_capacity[cur_nodes==0] = 0
        used_capacity = used_capacity + self.demand[cur_nodes]
        # update capacity_mask
        remaining_capacity = self.capacity - used_capacity # (n_ants,)
        remaining_capacity_repeat = remaining_capacity.unsqueeze(-1).repeat(1, self.problem_size) # (n_ants, p_size)
        demand_repeat = self.demand.unsqueeze(0).repeat(self.n_ants, 1) # (n_ants, p_size)
        capacity_mask[demand_repeat > remaining_capacity_repeat] = 0
        
        return used_capacity, capacity_mask
    
    def check_done(self, visit_mask, actions):
        return (visit_mask[:, 1:] == 0).all() and (actions == 0).all()
\end{lstlisting}

\paragraph{Manually Designed Prompts for TSP and CVRP in Existing Works.}

In prior works on evolutionary code search, prompt fragments for calling LLMs must be designed in accordance with the employed algorithmic framework, rather than the COP at hand.
In Table \ref{tab:aco-prompts}, we compare these components when ACO is employed for TSP vs. CVRP.

\begin{table}[h]
    \centering
    \footnotesize
    \caption{Prompt fragments used in ReEvo \citep{ye2024reevo} and MCTS-AHD \citep{zheng2025monte} under the ACO framework. }
    \subcaption*{TSP}
    \begin{tabular}{p{0.149\textwidth}|p{0.85\textwidth}}
        \hline
        \multicolumn{1}{c|}{Prompt fragment} & \multicolumn{1}{c}{Specification} \\\hline
        Problem description & Solving Traveling Salesman Problem (TSP) via stochastic solution sampling following ``heuristics''. TSP requires finding the shortest path that visits all given nodes and returns to the starting node. \\\hline
        Heuristic description & The `heuristics' function takes as input a distance matrix, and returns prior indicators of how promising it is to include each edge in a solution. The return is of the same shape as the input. \\\hline
        Function signature &
        \begin{pythontable}
def heuristics(distance_matrix: np.ndarray) -> np.ndarray:
\end{pythontable} \\
        \hline
    \end{tabular}
    \subcaption*{CVRP}
    \begin{tabular}{p{0.149\textwidth}|p{0.85\textwidth}}
        \hline
        \multicolumn{1}{c|}{Prompt fragment} & \multicolumn{1}{c}{Specification} \\\hline
        Problem description & Solving Capacitated Vehicle Routing Problem (CVRP) via stochastic solution sampling. CVRP requires finding the shortest path that visits all given nodes and returns to the starting node. Each node has a demand and each vehicle has a capacity. The total demand of the nodes visited by a vehicle cannot exceed the vehicle capacity. When the total demand exceeds the vehicle capacity, the vehicle must return to the starting node. \\\hline
        Heuristic description & The `heuristics' function takes as input a distance matrix (shape: n by n), Euclidean coordinates of nodes (shape: n by 2), a vector of customer demands (shape: n), and the integer capacity of vehicle capacity. It returns prior indicators of how promising it is to include each edge in a solution. The return is of the same shape as the distance matrix. The depot node is indexed by 0. \\\hline
        Function signature &
        \begin{pythontable}
def heuristics(distance_matrix: np.ndarray, coordinates: np.ndarray, demands: np.ndarray, capacity: int) -> np.ndarray:
\end{pythontable} \\
        \hline
    \end{tabular}
    \label{tab:aco-prompts}
\end{table}

\section{Additional Experiments and Discussions}\label{app:exp}

\subsection{Complete Implementation Details}\label{subapp:imp}

All experiments were conducted under Ubuntu 20.04 on a Linux virtual machine equipped with NVIDIA GeForce RTX 3050 Ti GPU and 12th Gen Intel(R) Core(TM) i7-12700H CPU @2.3GHz.
Code for our implementation in Python 3.10 is included in the supplementary material.

We adopt the experimental setups from MCTS-AHD \citep{zheng2025monte}, the state-of-the-art evolutionary code search, to better gauge the efficacy of ReFrame in solving COPs.
For the evaluation datasets, we use their publicly available data\footnote{\url{https://github.com/zz1358m/MCTS-AHD-master/tree/main}} during both training and testing for all considered COPs.

\paragraph{How the Initialization Prompt Is Split.}
Section \ref{subsec:init} generates a transformation with three separate LLM calls (candidate descriptions, then $(f_j,g_j)$, then the code template) rather than asking for everything at once.
Table \ref{tab:init-mode} compares that choice against two alternatives on CVRP: asking for all $M_0$ candidates and their code in a single call, and asking for one candidate per call with the previously generated descriptions supplied as context to avoid repetition.
Merging everything into one call is clearly worse, mostly because the code it emits frequently refers to fields that its own description did not define.
The sequential variant matches the default at roughly $M_0$ times the prompt tokens, so we use the default.

\begin{table}[h]
    \centering
    \footnotesize
    \caption{Initialization strategies, on CVRP ($n=50$, $C=50$), three runs each.
    ``Valid at init.'' is the fraction of the $M_0=10$ candidates whose transformation yields at least $\ell$ heuristics feasible on every training instance.}
    \begin{tabular}{l|cc}
        \thickhline
        Initialization strategy & Valid at init. & Final Obj. $\downarrow$ \\\hline
        Three calls (default) & 63\% & \textbf{9.826} \\
        One call for all candidates & 41\% & 10.284 \\
        One call per candidate, with context & 65\% & 9.841 \\
        \thickhline
    \end{tabular}
    \label{tab:init-mode}
\end{table}

\paragraph{ReFrame Settings.}\label{par:reframe-configs}

We set $M=3$, $M_0=10$, and $\ell=3$.
Prompts for transformation generation and refinement are specified in Figures \ref{fig:transform-init} and \ref{fig:transform-refine}, respectively.
The running time of each heuristic on the evaluation dataset for any COP is limited to 60 seconds.
We use EoH \citep{liu2024evolution} as the default evolutionary code search, in which we use only two variation operators, one for crossover and the other for mutation, instead of five as in the original work (see prompt specifications and our justifications in \nameref{par:eoh-configs}). We set $T$, the number of generations in EoH context, to 3.
Additionally, during population management at early stages of evolution, we do not discard heuristics with identical objective values if they are from different transformations. This ensures every transformation has sufficient heuristics (at least $\ell$) for obtaining a valid score.
Unless otherwise specified, GPT-4o-mini with temperature fixed at 1 (which is also the API default) is employed as the designer LLM for generating both transformations and heuristics, with each run of ReFrame repeated five times and we report the mean and standard deviation of the performance of $h^*$, i.e., for each run, the heuristic with the best training fitness is the one evaluated on the test instances.
Both designer models are accessed through the OpenAI API, where GPT-4o-mini and o3-mini denote the snapshots \texttt{gpt-4o-mini-2024-07-18} and \texttt{o3-mini-2025-01-31}, the only dated snapshots released for either model; o3-mini does not expose a temperature setting.

Because no handcrafted skeleton enforces feasibility, the feasibility predicate of Section \ref{subsec:residual} is what keeps the generated heuristics and transformations honest.
That is, during fitness evaluation, we check the solution of each instance as follows:
\begin{itemize}
    \item \textbf{TSP.} All nodes must be visited exactly once.
    \item \textbf{CVRP.} (i) Each customer from a sub-route must be visited exactly once; (ii) sum of demands from customers served by a sub-route must not exceed the vehicle capacity; (iii) all customers must be visited exactly once.
    \item \textbf{BPP.} All items must be packed in one of the bins without exceeding the capacity of any bin.
    \item \textbf{OBPP.} The selected bin must have sufficient capacity for packing the current item.
    \item \textbf{KP.} All selected items must be unique and their total weight must not exceed the knapsack capacity.
    \item \textbf{MKP.} All selected items across all knapsacks must be unique and the total weight of the items in any knapsack must not exceed its capacity.
    \item \textbf{Last-mile routing.} Every stop must be served within its time window, and the load of each route must not exceed the vehicle capacity.
\end{itemize}

\paragraph{Provenance of Baseline Numbers.}
As stated in their captions, the baseline rows of Tables \ref{tab:main-sbs}, \ref{tab:main-aco}, and \ref{tab:tsp-sbs} are reproduced from the literature \citep{zheng2025monte,ye2024reevo,duflo2019gp} rather than rerun by us, and the EoH* and ReEvo* reference rows that reappear in Tables \ref{tab:abl-eps}, \ref{tab:ablation-reduc}, \ref{tab:abl-refine}, and \ref{tab:abl-llm} repeat those numbers.
According to \citet{zheng2025monte}, the rows of the LLM-based searches in Tables \ref{tab:main-sbs} and \ref{tab:main-aco} were obtained with the same GPT-4o-mini snapshot at temperature $1$ and a budget of $1{,}000$ evaluations ($2{,}000$ for OBPP), and are averages over three runs, against five for ours; they also use each method's own variation operators and different hardware.
Differences against these rows should therefore be read as indicative rather than as a controlled comparison, and we report no significance tests against them.
The comparisons that hold the protocol fixed are those among our own runs, namely the ablations of Section \ref{subsec:ablations} and Appendix \ref{subapp:res}.
The test sets are those of \citet{zheng2025monte}, with $1{,}000$ instances per setting for TSP and KP under solution construction, $64$ per setting under ant colony optimization, and five per setting for OBPP.

\begin{figure}[h]
    \centering
    \begin{subfigure}[h]{0.98\textwidth}
        \centering
        \footnotesize
        \begin{tcolorbox}[width=\textwidth, title={Prompt for Candidate Transformation Initialization}, fonttitle=\bfseries\footnotesize]
        Problem A: \textcolor{mauve}{[Problem Description]}

        \hphantom\\

        I want to transform Problem A into another problem, Problem B, that can be solved efficiently while still providing near-optimal solutions to Problem A. Please help me devise \textcolor{purple}{$M_0$} different Problem B's. Describe each Problem B in a sentence or two (without mentioning Problem A) and enclose it inside a double brace as follows:

        \hphantom\\

        $\{\{$Problem B1 involves ...$\}\}$\\
        $\{\{$Problem B2 involves ...$\}\}$\\
        ...

        \hphantom\\
        
        Do not give additional explanations.
        \end{tcolorbox}
    \end{subfigure}\\
    \begin{subfigure}[h]{0.41\textwidth}
        \centering
        \footnotesize
        \begin{tcolorbox}[width=\textwidth, title={Prompt for Generating Transformation Functions}, fonttitle=\bfseries\footnotesize]
        Problem A: \textcolor{mauve}{[Problem A Description]}

        \hphantom\\

        Problem B: \textcolor{mauve}{[Problem B Description]}

        \hphantom\\

        Implement 2 Python functions for transforming Problem A into Problem B using the following templates:

        \hphantom\\

        \textcolor{dkgreen}{[Transformation Template]}

        \hphantom\\
        
        Only provide me the code without any further explanations.
        \end{tcolorbox}
    \end{subfigure}%
    ~
    \begin{subfigure}[h]{0.6\textwidth}
        \centering
        \footnotesize
        \begin{tcolorbox}[width=\textwidth, title={Prompt for Code Template Generation}, fonttitle=\bfseries\footnotesize]
        I have the following code for transforming a Problem A into a simplified Problem B and vice versa.

        \hphantom\\

        Code:\\
        \textcolor{dkgreen}{[Transformation Functions]}

        \hphantom\\

        Using this information, fill in the blanks of the following Python function template.

        \hphantom\\

        Code template:\\
        \textcolor{dkgreen}{[Heuristic Template]}

        \hphantom\\

        First, determine $<$INPUT\_B$>$ from output of `convert\_input\_A\_to\_B()'. Then, determine $<$SOLUTION\_B$>$ from `solution\_B' variable in `convert\_solution\_B\_to\_A()'.
        Finally, complete the docstring at $<$ARGS$>$ and $<$RETURNS$>$ with as detailed type hints as possible. Do not attempt to solve the problem directly and do not give additional explanations.
        \end{tcolorbox}
    \end{subfigure}
    \begin{minipage}{.55\textwidth}
    \begin{lstlisting}[captionpos=b]
import numpy as np
from typing import Tuple

def convert_input_A_to_B(*input_A):
    ''' Convert input of Problem A into input of Problem B
    Args:
    [ARGS]

    Returns:
    input_B: A tuple storing the corresponding input of Problem B.
    '''

    # Placeholder (replace with your actual implementation)
    input_B = ...

    return input_B


def convert_solution_B_to_A(solution_B):
    ''' Convert solution of Problem B into solution of Problem A
    Args:
    solution_B: The output of Problem B.

    Returns:
    [RETURN]
    '''

    # Placeholder (replace with your actual implementation)
    (*@\aftergroup\speciallstcolor@*)[PLACEHOLDER](*@\aftergroup\endspeciallstcolor@*)
    \end{lstlisting}
    \end{minipage}\hfill
    \begin{minipage}{.4\textwidth}
    \begin{lstlisting}[captionpos=b]
from typing import Tuple

def solve_B(<INPUT_B>):
    '''
    Args:
    <ARGS>

    Returns:
    <RETURNS>
    '''

    return <SOLUTION_B>
    \end{lstlisting}
    \end{minipage}
    \caption{Prompts used for candidate transformation generation in ReFrame as described in Section \ref{subsec:init}. The chronological order for LLM prompting is (top) $\rightarrow$ (center left) $\rightarrow$ (center right). The (bottom left) code snippet is the ``Transformation Template'', where [ARGS], [RETURN], [PLACEHOLDER] are COP-specific and detailed in Table \ref{tab:cop-transform-prompts}. The (bottom right) code snippet is the ``Heuristic Template''.}
    \label{fig:transform-init}
\end{figure}

\begin{figure}[h]
    \centering
    \begin{subfigure}[h]{0.9\textwidth}
        \centering
        \footnotesize
        \begin{tcolorbox}[width=\textwidth, title={Prompt for Transformation Refinement}, fonttitle=\bfseries\footnotesize]
        Problem A: \textcolor{mauve}{[Problem A Description]}

        \hphantom\\

        I want to transform Problem A into another problem, Problem B, that can be solved efficiently while still providing near-optimal solutions to Problem A. I have one option for Problem B as follows:

        \hphantom\\

        Problem description: \textcolor{mauve}{[Problem B Description]}

        \hphantom\\

        Please help me modify the following code for transforming Problem A to Problem B and vice versa while remaining as efficient as possible.

        \hphantom\\

        Code:\\
        \textcolor{dkgreen}{[Transformation Functions]}

        \hphantom\\

        Do not give additional explanations.
        \end{tcolorbox}
    \end{subfigure}
    \caption{Prompts used for transformation refinement in ReFrame as described in Section \ref{subsec:refine}.}
    \label{fig:transform-refine}
\end{figure}

\clearpage
\paragraph{EoH Settings.}\label{par:eoh-configs}
Following \citet{zheng2025monte}, the number of generations in EoH is set to 20 and the population size $\mu$ is set to 20 for CVRP, BPP, OBPP, MKP and 10 for TSP and KP.
EoH utilizes five variation operators in total, two for crossover (E1 and E2) and three for mutation (M1, M2, M3).
ReFrame only uses E2 and M1 from EoH (see Figure \ref{fig:eoh-prompt-design}, bottom) since we actually observed reduced optimization performance when either E1, M2, or M3 is included (and significant increase in runtime and API cost).
In particular, we notice the heuristics generated by E1 are often erroneous (due to e.g., code errors or returning invalid solutions).
We attribute this behavior to the fact that E1 prompts the designer LLM to generate a completely new heuristic from the provided ones, which might not be well-suited for multi-problem evolutionary code search within ReFrame that already enables ample exploration of novel heuristics.
This two-operator choice was made in preliminary runs and then fixed for every problem and every table; a paired two- versus five-operator comparison across all problems is not part of the reported experiments.
Note that the EoH* rows we reproduce from the literature use that work's own operator configuration, so the pairs EoH* versus ReFrame[EoH] differ in operator set as well as in specification.

\begin{figure}[h]
    \centering
    \begin{subfigure}[h]{0.8\textwidth}
        \centering
        \footnotesize
        \begin{tcolorbox}[width=\textwidth, title={Prompt for Initialization}, fonttitle=\bfseries\footnotesize]
        \textcolor{mauve}{[Problem Description]}

        \hphantom\\

        I need help design a novel efficient algorithm to solve the problem. First, describe your algorithm and main steps in one sentence. The description must be inside a brace. Next, implement it in Python using the following template:

        \hphantom\\

        \textcolor{dkgreen}{[Code Template]}

        \hphantom\\
        
        Do not give additional explanations.
        \end{tcolorbox}
    \end{subfigure}\\
    \begin{subfigure}[h]{0.54\textwidth}
        \centering
        \footnotesize
        \begin{tcolorbox}[width=\textwidth, title={Prompt for Crossover/Exploration}, fonttitle=\bfseries\footnotesize]
        \textcolor{mauve}{[Problem Description]}

        \hphantom\\

        I have 2 existing algorithms with their codes as follows:\\
        No. 1 algorithm and the corresponding code are:\\
        \textcolor{blue}{[Algorithm 1 Description]}\\
        \textcolor{dkgreen}{[Code 1]}

        \hphantom\\

        No. 2 algorithm and the corresponding code are:\\
        \textcolor{blue}{[Algorithm 2 Description]}\\
        \textcolor{dkgreen}{[Code 2]}

        \hphantom\\

        Please help me create a new algorithm that has a totally different form from the given ones but can be motivated from them.
        First, identify the common backbone idea in the provided algorithms. Secondly, based on the backbone idea describe your new algorithm in one sentence. The description must be inside a brace. Thirdly, implement it in Python using the following template:

        \hphantom\\

        \textcolor{dkgreen}{[Code Template]}

        \hphantom\\
        
        Do not give additional explanations.
        \end{tcolorbox}
    \end{subfigure}%
    ~
    \begin{subfigure}[h]{0.45\textwidth}
        \centering
        \footnotesize
        \begin{tcolorbox}[width=\textwidth, title={Prompt for Mutation/Modification}, fonttitle=\bfseries\footnotesize]
        \textcolor{mauve}{[Problem Description]}

        \hphantom\\

        I have one algorithm with its code as follows.\\
        Algorithm description:  \textcolor{blue}{[Algorithm Description]}\\
        Code: \textcolor{dkgreen}{[Code]}

        \hphantom\\

        Please help me create a new algorithm that has a different form but can be a modified version of the provided algorithm.
        First, describe your new algorithm and main steps in one sentence. The description must be inside a brace. Next, implement it in Python using the following template:

        \hphantom\\

        \textcolor{dkgreen}{[Code Template]}

        \hphantom\\
        
        Do not give additional explanations.
        \end{tcolorbox}
    \end{subfigure}
    \caption{Prompts used for initialization, exploration, and modification in EoH. ``Problem Description'' and ``Code Template'' are with respect to $B$ from the LLM-generated transformation (see Figure \ref{fig:learned-transform-mkp} for an example).}
    \label{fig:eoh-prompt-design}
\end{figure}

\paragraph{MEoH Settings.}

MEoH \citep{yao2024multi} extends EoH to additionally consider runtime during fitness evaluation via the proposed dominance-dissimilarity mechanism for multi-objective parent selection and population management.
We similarly use two variation operators as detailed in \nameref{par:eoh-configs}.
Importantly, each transformation now records two scores, one with respect to the objective value and the other with respect to runtime.
The latter is defined as the average runtime of its top-$\ell$ associated heuristics with best objective values.
For stagnation tracking, if neither score improves after $T$, then the transformation refinement step is invoked for the transformation.

\paragraph{ReEvo Settings.}
ReEvo \citep{ye2024reevo} incorporates reflections into the evolutionary search by prompting the designer LLM to analyze and revise previously generated heuristics.
We make the following changes to ReEvo.
During parent selection, offspring allocation is similarly applied to maintain the number of generated offspring heuristics from the two crossover and mutation operators.
Short- and long-term reflections are performed for each transformation.
For short-term reflection, the problem description is with respect to $B_j$. Importantly, in accordance with our proposed multi-problem evolutionary code search in ReFrame, the two provided heuristics can be from $B_{j'}, j'\neq j$.

\paragraph{COP-Specific Prompts.}

Tables \ref{tab:cop-prompts} and \ref{tab:cop-transform-prompts} respectively list the problem descriptions and transformation templates used in prompts.
To facilitate the generation of valid transformations that can generalize to OOD instances (i.e., instances with smaller or larger sizes than what originally encountered during training), when specifying problem descriptions and transformation templates, we ensure all COP parameters are abstracted, such as `N' instead of the actual number of nodes in training instances.
Altogether, for each new problem the user of ReFrame writes the problem statement (Table \ref{tab:cop-prompts}), the typed input and solution formats that fill the transformation template (the [ARGS], [RETURN], and [PLACEHOLDER] fragments of Table \ref{tab:cop-transform-prompts}), and the feasibility predicate listed above, in addition to the instance loading and the objective function that every framework needs.
The line count of Section \ref{subsec:effort} covers the problem statement, the template fragments (five to eight lines for each problem of Table \ref{tab:cop-transform-prompts}), and the feasibility predicate; instance loading and the objective function are excluded for every method.

\begin{table}[h]
    \centering
    \footnotesize
    \caption{Problem descriptions used in prompts. }
    \begin{tabular}{l|p{0.9\textwidth}}
        \hline
        \multicolumn{1}{c|}{COP} & \multicolumn{1}{c}{Problem description} \\\hline
        TSP & Given a set of N nodes with their 2D coordinates, the problem involves finding the shortest route that visits each node exactly once and returns to the starting node. \\\hline
        CVRP & Given a set of N customers and a fleet of vehicles with limited capacity, the problem involves finding a corresponding set of optimal routes to deliver goods to all customers. \\\hline
        BPP & Given a set of N items with different sizes and some bins each with fixed capacity, the problem involves placing each item inside one of the bins in a way that minimizes the number of bins used without exceeding the bin capacity. \\\hline
        OBPP & Given an item with certain size and a set of M bins each with finite capacity, the problem involves finding a priority score for each bin. The bin with the highest priority score will be selected for inserting the item. \\\hline
        KP & Given a set of N items with weights and values, the problem involves selecting a subset of items that maximizes the total value without exceeding the knapsack's weight capacity. \\\hline
        MKP & Given a set of N items with values and M-dimensional weights, the problem involves selecting a subset of items to maximize the total value without exceeding the multi-dimensional maximum weight constraints. \\
        \hline
    \end{tabular}
    \label{tab:cop-prompts}
\end{table}

\begin{footnotesize}
\begin{longtable}{l|p{0.8\textwidth}}
    \caption{COP-specific components for transformation templates.}\\
        \hline
        \multicolumn{1}{c|}{Component} & \multicolumn{1}{c}{Specification} \\\hline
        \multicolumn{2}{c}{\cellcolor{green!25}TSP}\\\hline
        ARGS & \begin{pythontable}
'''
coord_matrix (np.ndarray): A Nx2 matrix storing the 2D coordinates of the nodes.
distance_matrix (np.ndarray): A NxN matrix where the entry at i-th row and j-th column (or vice versa) stores the Euclidean distance between nodes i and j.
'''
\end{pythontable} \\[7ex]\hline
        RETURN & \begin{pythontable}
'''
route: A Numpy 1D array of length N storing the unique node IDs to visit in order.
'''
\end{pythontable} \\[5ex]\hline
        PLACEHOLDER & \begin{pythontable}
route = ...

return route
\end{pythontable} \\[5ex]\hline
        \multicolumn{2}{c}{\cellcolor{green!25}CVRP}\\\hline
        ARGS & \begin{pythontable}
'''
coord_matrix (np.ndarray): A (N+1)-by-2 matrix storing the Euclidean coordinates of the depot (first row) and the customers. 
distance_matrix (np.ndarray): A (N+1)-by-(N+1) distance matrix.
demands (np.ndarray): An array of length N+1 storing the customer demands, where the first entry is 0 (placeholder for the depot).
capacity (int): The capacity of each vehicle for satisfying the customer demands.
'''
\end{pythontable} \\[12ex]\hline
        RETURN & \begin{pythontable}
'''
routes (List[List[int]]): A list of routes; each route is represented as a list of unique customer indices (1 to N) to visit in order, subject to the capacity constraint.
'''
\end{pythontable} \\[7ex]\hline
        PLACEHOLDER & \begin{pythontable}
routes = []
...

return routes
\end{pythontable} \\[7ex]\hline
        \multicolumn{2}{c}{\cellcolor{green!25}BPP}\\\hline
        ARGS & \begin{pythontable}
'''
items (np.ndarray): Array of length N storing the item sizes to be considered in exact order.
bins (np.ndarray): Array of capacities for each bin.
'''
\end{pythontable} \\[7ex]\hline
        RETURN & \begin{pythontable}
'''
packed_bins (np.ndarray): Array of remaining capacities for each bin after packing all items.
'''
\end{pythontable} \\[5ex]\hline
        PLACEHOLDER & \begin{pythontable}
packed_bins = ...
...

return packed_bins
\end{pythontable} \\[7ex]\hline
        \multicolumn{2}{c}{\cellcolor{green!25}OBPP}\\\hline
        ARGS & \begin{pythontable}
'''
item_size (float): Size of the item to be added to one of the bins.
bin_caps (np.ndarray): Array of length M storing capacities of each bin.
'''
\end{pythontable} \\[7ex]\hline
        RETURN & \begin{pythontable}
'''
scores (np.ndarray): Array of priority scores for the bins.
'''
\end{pythontable} \\[5ex]\hline
        PLACEHOLDER & \begin{pythontable}
scores = ...
...

return scores
\end{pythontable} \\[7ex]\hline
        \multicolumn{2}{c}{\cellcolor{green!25}KP}\\\hline
        ARGS & \begin{pythontable}
'''
weights (np.ndarray): A 1D float array of length {problem_size} storing the item weights.
values (np.ndarray): A 1D float array of length {problem_size} storing the associated item values.
capacity (float): The weight capacity of the knapsack.
'''
\end{pythontable} \\[10ex]\hline
        RETURN & \begin{pythontable}
'''
items: A list storing the indices of selected items subject to the capacity constraint.
'''
\end{pythontable} \\[5ex]\hline
        PLACEHOLDER & \begin{pythontable}
items = []
...

return items
\end{pythontable} \\[7ex]\hline
        \multicolumn{2}{c}{\cellcolor{green!25}MKP}\\\hline
        ARGS & \begin{pythontable}
'''
values (np.ndarray): A 1D float array of length N storing the item values.
weights (np.ndarray): A (M x N) float matrix storing the multi-dimensional weights, where each row is associated with a constraint.
constraints (np.ndarray): A 1D float array of length M storing weight constraints.
'''
\end{pythontable} \\[10ex]\hline
        RETURN & \begin{pythontable}
'''
items: A list storing the indices of selected items subject to the weight constraints.
'''
\end{pythontable} \\[5ex]\hline
        PLACEHOLDER & \begin{pythontable}
items = []
...

return items
\end{pythontable} \\[7ex]
        \hline
    \label{tab:cop-transform-prompts}
\end{longtable}
\end{footnotesize}

\subsection{Additional Results}\label{subapp:res}


\paragraph{Where the Multi-Problem Gain Comes From.}
Multi-problem evolutionary code search varies two things at the same time, namely how many transformations are retained and whether a parent heuristic may be drawn from a target problem other than the one the offspring is written for.
Table \ref{tab:ablation-reduc} separates them.
The second row sets $M=1$, so that the search reduces to typical evolutionary code search on a single synthesized transformation under the same evaluation budget.
The third row keeps $M=3$ but partitions the population by target problem, so that each transformation is evolved in its own slice and no parent is ever drawn across slices.
The fourth row is ReFrame as described in Section \ref{subsec:multiproblem}, in which a parent may be any heuristic in $P$.

\begin{figure}[h]
  \centering
  \includegraphics[width=0.46\linewidth]{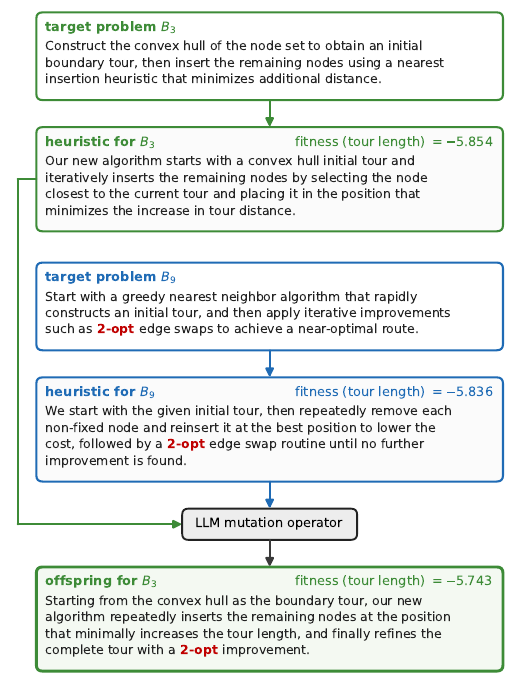}
  \caption{Why the target problems are evolved together.
  The parent heuristic (blue) was written for a different target problem than the one being extended, and describes a $2$-opt edge exchange.
  Applying the mutation operator across target problems carries that idea into the population for $B_3$ (green), where nothing in the local population suggested it, and the offspring is better than any of its siblings.
  Ideas move in this direction only because parents are drawn from the whole population rather than from one target problem's slice of it.}
  \label{fig:demo}
\end{figure}

Partitioning is worse than $M=1$ on every one of the eight settings, and both are well behind ReFrame, for instance on CVRP at $n=50$, where $M=1$ gives $10.327$, the partitioned variant $10.537$, and ReFrame $9.826$.
Retaining three transformations therefore buys nothing on its own.
We attribute this to the fact that the offspring budget of a generation is split three ways, so that each population is explored roughly a third as deeply as in the $M=1$ case while no algorithmic idea is allowed to move between them.
What carries the framework is thus the cross-problem parent selection rather than the number of transformations retained, which is the direct evidence for the design decision in Section \ref{subsec:multiproblem}.
This evidence is at the level of outcomes, and Figure \ref{fig:demo} traces a single transfer; a lineage-level analysis of how often cross-problem parents are selected and how often their offspring survive would quantify the mechanism more directly and is left for future work.

\begin{table}[h]
    \centering
    \footnotesize
    \caption{Ablation of the proposed multi-problem evolutionary code search. ``ReFrame ($M=3$)'' is ReFrame reported in earlier results. Results from EoH in Table \ref{tab:main-aco} are used as references.}
    \begin{tabular}{l|cc|cc|cc|cc}
        \thickhline
        \multirow{3}{*}{\backslashbox{Method}{\strut Problem\\ setting}} & \multicolumn{2}{c|}{TSP (Obj. $\downarrow$)} & \multicolumn{2}{c|}{CVRP (Obj. $\downarrow$)} & \multicolumn{2}{c|}{MKP (Obj. $\uparrow$)} & \multicolumn{2}{c}{BPP (Obj. $\downarrow$)} \\\cline{2-9}
        & \underline{$n$=50} & $n$=100 & \makecell{\underline{$n$=50}\\ \underline{$C$=50}} & \makecell{$n$=100\\$C$=50} & \makecell{\underline{$n$=100}\\\underline{$m$=5}} & \makecell{$n$=200\\$m$=5} & \makecell{\underline{$n$=500}\\\underline{$W$=150}} & \makecell{$n$=1,000\\$W$=150} \\\hline
        EoH* & 5.828 & 8.263 & \textbf{9.359} & \textbf{15.681} & 23.139 & 41.994 & 204.646 & 408.599 \\\hline
        ReFrame ($M=1$) & 5.931 & 8.479 & 10.327 & 16.252 & 22.925 & 41.569 & 205.983 & 411.428 \\
        ReFrame ($M=3$, without multi-problem evolutionary code search) & 5.943 & 8.602 & 10.537 & 16.985 & 22.916 & 41.497 & 206.013 & 412.220 \\
        ReFrame ($M=3$) & \textbf{5.819} & \textbf{8.039} & 9.826 & 15.726 & \textbf{23.164} & \textbf{42.682} & \textbf{203.344} & \textbf{405.359} \\
        \thickhline
    \end{tabular}
    \label{tab:ablation-reduc}
\end{table}

\clearpage
\paragraph{Sensitivity to Initial Transformations.}
We investigate the sensitivity of ReFrame performance to the quality of the initial pool of transformations. Table \ref{tab:sensi} compares the average test performance of all heuristics in the initial generation (second column) to the test performance of the best heuristic in the final generation (last column).
Even though the quality of the initial transformations differs across runs, the final performance of ReFrame remains consistent.

\begin{table}[h]
    \footnotesize
    \centering
    \caption{ReFrame performance on TSP ($n=50$) across five independent runs (lower values are better).}
    \begin{tabular}{c|c|c}
        \thickhline
        Run & Quality of initial transformations & Final performance \\\hline
        1 & 6.831 & 5.784 \\
        2 & 6.378 & 5.761 \\
        3 & 6.494 & 5.775 \\
        4 & 6.796 & 5.770 \\
        5 & 6.592 & 5.766 \\
        \thickhline
    \end{tabular}
    \label{tab:sensi}
\end{table}

\paragraph{Real-World Instances.}
Table \ref{tab:tsp-sbs} reports optimality gaps on TSPLib \citep{reinelt1991tsplib}, whose instances have geometry unlike the uniform-random training data.
The heuristic designed here is best on every instance, and holds a single-digit gap on instances above $1{,}500$ nodes, where the asterisked searches were not run and where Christofides' approximation algorithm is the strongest of the classical baselines.
We read this as a generalization result about the heuristic rather than about the framework, since the transformation selected for TSP produces a construction rule that does not depend on the instance distribution.
That is a property of that particular transformation and not something the method guarantees.

\begin{table}[h]
    \centering
    \footnotesize
    \caption{Optimality gap (\%, lower is better) on TSPLib against published optima.
    The number in each instance name is its node count, and instances above $1{,}500$ nodes are highlighted.
    Searches marked with an asterisk use a handcrafted solution-construction skeleton.
    Baseline numbers are as reported by \citet{duflo2019gp,ye2024reevo,zheng2025monte}.
    Christofides' algorithm \citep{christofides1976worst} is a $1.5$-approximation algorithm rather than a metaheuristic.
    Our column is the best heuristic across five runs, evaluated three times per instance because it randomizes its starting node, reported as mean\,$\pm$\,std over those repetitions.
    The best baseline is shaded and the overall best is bold.}
    \begin{tabular}{l|ccccc|ccc|c}
        \thickhline
        \makecell[c]{TSPLib\\ instance} & Christofides \citeyear{christofides1976worst} & Greedy \citeyear{brecklinghaus2015approximation} & \makecell[c]{Nearest\\ insertion} & \makecell[c]{Nearest\\ neighbor} \citeyear{rosenkrantz1977analysis} & GPHH-best \citeyear{duflo2019gp} & EoH* & ReEvo* & MCTS-AHD* & ReFrame \\\hline
        ts225 & 5.67 & \cellcolor{gray!25}5.38 & 19.93 & 16.82 & 7.71 & 5.57 & 6.56 & 10.84 & \textbf{2.29 $\pm$ 0.21} \\
        rat99 & \cellcolor{gray!25}9.43 & 22.30 & 21.05 & 21.79 & 14.09 & 18.78 & 12.41 & 10.46 & \textbf{3.47 $\pm$ 0.08} \\
        \cellcolor{green!25}rl1889 & \cellcolor{gray!25}7.60 & 19.44 & 24.34 & 23.74 & 21.09 & - & 17.5 & - & \textbf{6.87 $\pm$ 0.61} \\
        \cellcolor{green!25}u1817 & \cellcolor{gray!25}14.15 & 19.78 & 24.07 & 22.20 & 21.21 & - & 16.6 & - & \textbf{6.42 $\pm$ 0.16} \\
        \cellcolor{green!25}d1655 & \cellcolor{gray!25}12.65 & 16.31 & 21.35 & 23.86 & 18.69 & - & 17.5 & - & \textbf{7.10 $\pm$ 0.34} \\
        bier127 & 13.03 & 19.50 & 23.05 & 23.25 & 15.64 & 14.05 & 10.79 & \cellcolor{gray!25}7.56 & \textbf{2.32 $\pm$ 0.38} \\
        lin318 & \cellcolor{gray!25}13.80 & 18.75 & 24.44 & 25.78 & 14.30 & 14.03 & 16.63 & 14.07 & \textbf{5.39 $\pm$ 0.17} \\
        eil51 & 15.18 & 13.03 & 16.14 & 31.96 & 10.20 & 8.37 & \cellcolor{gray!25}6.47 & 15.98 & \textbf{2.29 $\pm$ 0.48} \\
        d493 & \cellcolor{gray!25}9.52 & 16.68 & 20.39 & 24.00 & 15.58 & 12.41 & 13.43 & 11.73 & \textbf{3.83 $\pm$ 0.28} \\
        kroB100 & \cellcolor{gray!25}9.82 & 16.59 & 21.53 & 26.26 & 14.06 & 13.46 & 12.20 & 11.43 & \textbf{2.12 $\pm$ 0.84} \\
        kroC100 & 9.08 & 12.94 & 24.25 & 25.76 & 16.22 & 16.85 & 15.88 & \cellcolor{gray!25}8.27 & \textbf{3.64 $\pm$ 0.24} \\
        ch130 & 10.09 & 28.40 & 19.21 & 25.66 & 14.77 & 12.26 & \cellcolor{gray!25}9.40 & 10.18 & \textbf{4.51 $\pm$ 0.69} \\
        pr299 & \cellcolor{gray!25}11.23 & 31.42 & 25.05 & 31.42 & 18.24 & 23.58 & 20.63 & \cellcolor{gray!25}11.23 & \textbf{5.45 $\pm$ 0.33} \\
        fl417 & 15.57 & 12.64 & 25.52 & 32.42 & 22.72 & 20.47 & 19.15 & \cellcolor{gray!25}10.20 & \textbf{3.43 $\pm$ 0.52} \\
        d657 & \cellcolor{gray!25}10.41 & 15.76 & 22.84 & 29.74 & 16.30 & - & 16.0 & - & \textbf{5.34 $\pm$ 0.61} \\
        kroA150 & 13.44 & 20.24 & 19.09 & 26.08 & 15.59 & 18.36 & 11.62 & \cellcolor{gray!25}10.08 & \textbf{3.62 $\pm$ 0.31} \\
        \cellcolor{green!25}fl1577 & \cellcolor{gray!25}8.84 & 15.60 & 24.17 & 25.01 & 17.60 & - & 12.1 & - & \textbf{3.17 $\pm$ 0.38} \\
        u724 & \cellcolor{gray!25}12.04 & 17.20 & 25.58 & 28.45 & 15.54 & - & 16.9 & - & \textbf{5.08 $\pm$ 0.38} \\
        pr264 & \cellcolor{gray!25}11.28 & 11.89 & 34.28 & 17.87 & 23.96 & 18.03 & 16.78 & 12.27 & \textbf{4.97 $\pm$ 1.05} \\
        pr226 & 14.17 & 21.44 & 28.02 & 24.65 & 15.51 & 19.90 & 18.02 & \cellcolor{gray!25}7.15 & \textbf{1.97 $\pm$ 1.13} \\
        pr439 & \cellcolor{gray!25}11.16 & 20.08 & 24.67 & 27.36 & 21.36 & 21.96 & 19.25 & 15.12 & \textbf{5.65 $\pm$ 0.50} \\
        \thickhline
    \end{tabular}
    \label{tab:tsp-sbs}
\end{table}



\clearpage
\paragraph{Cost of a Design Run.}
Table \ref{tab:cost} breaks a design run into its three stages.
Initialization is $19\%$ of wall-clock and about $8\%$ of LLM calls, and refinement $15\%$ and under $3\%$ of calls, so the evolutionary search dominates both, as it does in the methods we build on.
The overhead our framework adds over running the same search inside a handcrafted skeleton is therefore roughly a third of wall-clock and a tenth of LLM calls, and in return the skeleton itself no longer has to be written.

\begin{table}[h]
    \centering
    \footnotesize
    \caption{Cost of one design run with GPT-4o-mini, broken down by stage, and the inference time of the resulting heuristic.
    Wall-clock is dominated by fitness evaluation rather than by LLM latency.
    ``Calls'' counts LLM queries: $1+2M_0$ during initialization, two per offspring during evolution, and one per refinement.
    Money is at the GPT-4o-mini list price.
    With o3-mini as the designer the same runs cost $6$--$7\times$ more and take $10$--$20\%$ longer.
    Inference is the mean wall-clock of the final heuristic on one test instance at the largest size reported for that problem.}
    \begin{tabular}{l|rrr|r|rrr|r}
        \thickhline
        \multirow{2}{*}{Problem} & \multicolumn{3}{c|}{Wall-clock (min)} & \multirow{2}{*}{\makecell{Total\\(min)}} & \multirow{2}{*}{Calls} & \multirow{2}{*}{\makecell{Tokens\\(M)}} & \multirow{2}{*}{\makecell{Cost\\(USD)}} & \multirow{2}{*}{\makecell{Inference\\(ms/inst.)}} \\\cline{2-4}
        & init. & evol. & refine & & & & & \\\hline
        TSP  & 13.4 & 49.8 & \phantom{0}9.6 & \phantom{0}72.8 & 265 & 0.51 & 0.11 & \phantom{00}41 \\
        CVRP & 15.1 & 61.2 & 12.7 & \phantom{0}89.0 & 267 & 0.58 & 0.13 & \phantom{0}190 \\
        KP   & \phantom{0}9.8 & 31.5 & \phantom{0}7.2 & \phantom{0}48.5 & 264 & 0.44 & 0.09 & \phantom{000}6 \\
        MKP  & 11.2 & 38.4 & \phantom{0}8.9 & \phantom{0}58.5 & 265 & 0.47 & 0.10 & \phantom{00}22 \\
        BPP  & 12.6 & 44.7 & 10.3 & \phantom{0}67.6 & 266 & 0.49 & 0.10 & \phantom{00}58 \\
        OBPP & 10.4 & 35.9 & \phantom{0}8.1 & \phantom{0}54.4 & 266 & 0.46 & 0.10 & \phantom{00}74 \\
        Last-mile & 16.8 & 68.3 & 14.5 & \phantom{0}99.6 & 268 & 0.62 & 0.14 & \phantom{0}260 \\\hline
        \emph{share} & 19\% & 66\% & 15\% & & & & & \\
        \thickhline
    \end{tabular}
    \label{tab:cost}
\end{table}

\paragraph{Removing Transformation Refinement.}
Table \ref{tab:abl-refine} reruns the ant colony optimization comparison with refinement disabled.
Performance drops on every problem, and on CVRP and TSP the result falls behind the search it is built on.
The mechanism is the one described in Section \ref{subsec:refine}.
A transformation that is lucky early absorbs the sampling budget of Section \ref{subsec:multiproblem}, the population fills with heuristics written against it, and the search converges inside a representation that cannot express anything better.

\begin{table}[h]
    \centering
    \footnotesize
    \caption{Removing transformation refinement.
    EoH inside a handcrafted ant colony optimization skeleton, from Table \ref{tab:main-aco}, is shown for reference.}
    \begin{tabular}{l|cc|cc|cc|cc}
        \thickhline
        \multirow{3}{*}{\backslashbox[42mm]{Method}{\strut Problem\\ setting}} & \multicolumn{2}{c|}{TSP (Obj. $\downarrow$)} & \multicolumn{2}{c|}{CVRP (Obj. $\downarrow$)} & \multicolumn{2}{c|}{MKP (Obj. $\uparrow$)} & \multicolumn{2}{c}{BPP (Obj. $\downarrow$)} \\\cline{2-9}
        & \underline{$n$=50} & $n$=100 & \makecell{\underline{$n$=50}\\ \underline{$C$=50}} & \makecell{$n$=100\\$C$=50} & \makecell{\underline{$n$=100}\\\underline{$m$=5}} & \makecell{$n$=200\\$m$=5} & \makecell{\underline{$n$=500}\\\underline{$W$=150}} & \makecell{$n$=1,000\\$W$=150} \\\hline
        EoH* (reference) & 5.828 & 8.263 & \textbf{9.359} & \textbf{15.681} & 23.139 & 41.994 & 204.646 & 408.599 \\\hline
        ReFrame without refinement & 5.847 & 8.322 & 10.218 & 16.175 & 23.126 & 41.978 & 204.561 & 407.639 \\
        ReFrame & \textbf{5.819} & \textbf{8.039} & 9.826 & 15.726 & \textbf{23.164} & \textbf{42.682} & \textbf{203.344} & \textbf{405.359} \\
        \thickhline
    \end{tabular}
    \label{tab:abl-refine}
\end{table}

\paragraph{Changing the Designer LLM\@.}
Table \ref{tab:abl-llm} swaps GPT-4o-mini for o3-mini and changes nothing else.
OBPP improves in all settings, some of which even surpass the best baseline, and CVRP improves by $15.0\%$ in-distribution and $14.1\%$ out-of-distribution, moving from behind OR-Tools to level with it and to within $1.6\%$ of HGS.
Since the search, the budget, and the evaluator are identical across the two rows, the difference is attributable to the designer LLM; because the same model writes both the transformations and the heuristics, this table does not by itself separate the two, and the reference rows are reproduced numbers that were not rerun with o3-mini.

\begin{table}[h]
    \centering
    \footnotesize
    \caption{Changing only the designer LLM.
    Reference rows are truncated from Tables \ref{tab:main-sbs} and \ref{tab:main-aco}.}
    \begin{tabular}{l|cccccc}
        \thickhline
        \multicolumn{7}{c}{OBPP (\% gap to the bin lower bound $\downarrow$)} \\\hline
        \multicolumn{1}{c|}{$n$ (items)} & \underline{1k} & \underline{1k} & \underline{5k} & \underline{5k} & 10k & 10k \\
        \multicolumn{1}{c|}{$W$ (bin capacity)} & \underline{100} & \underline{500} & \underline{100} & \underline{500} & 100 & 500 \\\hline
        Best baseline & \textbf{2.45} & 0.25 & \textbf{1.06} & \textbf{0.25} & \textbf{0.74} & 0.21 \\
        EoH* & 2.69 & 0.25 & 1.63 & 0.53 & 1.47 & 0.45 \\\hline
        ReFrame (GPT-4o-mini) & 3.78 & 0.99 & 2.82 & 0.55 & 2.61 & 0.40 \\
        ReFrame (o3-mini) & 3.13 & \textbf{0.00} & 2.33 & 0.30 & 2.02 & \textbf{0.20} \\
        \thickhline
    \end{tabular}
    \hspace{2cm}
    \begin{tabular}{l|cc}
        \thickhline
        \multicolumn{3}{c}{CVRP (Obj. $\downarrow$)} \\\hline
        \multicolumn{1}{c|}{$n$ (nodes)} & \underline{50} & 100 \\
        \multicolumn{1}{c|}{$C$ (vehicle capacity)} & \underline{50} & 50 \\\hline
        HGS \citeyear{vidal2022hybrid}, 60\,s & \textbf{8.116} & \textbf{13.302} \\
        OR-Tools \citeyear{ortools_routing}, 60\,s & 8.314 & 13.948 \\
        DeepACO \citeyear{ye2023deepaco} & 8.888 & 14.932 \\
        EoH* & 9.359 & 15.681 \\\hline
        ReFrame (GPT-4o-mini) & 9.826 & 15.726 \\
        ReFrame (o3-mini) & 8.348 & 13.516 \\
        \thickhline
    \end{tabular}
    \label{tab:abl-llm}
\end{table}


\paragraph{How Many Transformations, and How Patiently.}
Table \ref{tab:abl-mt} sweeps $M$, the number of transformations evolved jointly, and $T$, the number of stagnant generations tolerated before a transformation is rewritten.
$M=1$ removes multi-problem search entirely and is clearly worse, $M=3$ captures nearly all of the available gain, and $M=5$ and $M=7$ add little while multiplying the initialization cost.
The behavior of $T$ is less obvious, since $T=1$ is worse than $T=3$ rather than better.
We attribute this to the fact that rewriting a transformation after a single stagnant generation discards representations whose heuristic space had not yet been explored, so that the search never finds out whether they were good.
Large $T$ approaches the no-refinement ablation above.
Patience is therefore a real parameter and not a formality, and $T=3$ is a compromise rather than a tuned optimum, used everywhere without per-problem tuning.

\begin{table}[h]
    \centering
    \footnotesize
    \caption{Sensitivity to the number of jointly evolved transformations $M$ and the refinement patience $T$.
    Mean over five runs.
    The setting used everywhere else in the paper is $M=3$, $T=3$.}
    \begin{tabular}{l|cc||l|cc}
        \thickhline
        \multirow{2}{*}{Setting} & \makecell{TSP\\ \underline{$n$=50}} & \makecell{CVRP\\ \underline{$n$=50}} & \multirow{2}{*}{Setting} & \makecell{TSP\\ \underline{$n$=50}} & \makecell{CVRP\\ \underline{$n$=50}} \\
        & Obj. $\downarrow$ & Obj. $\downarrow$ & & Obj. $\downarrow$ & Obj. $\downarrow$ \\\hline
        $M$=1 & 5.931 & 10.327 & $T$=1 & 5.884 & 10.043 \\
        $M$=3 & 5.819 & 9.826 & $T$=3 & \textbf{5.819} & \textbf{9.826} \\
        $M$=5 & \textbf{5.811} & \textbf{9.794} & $T$=5 & 5.828 & 9.882 \\
        $M$=7 & 5.822 & 9.851 & no refine & 5.847 & 10.218 \\
        \thickhline
    \end{tabular}
    \label{tab:abl-mt}
\end{table}

\begin{figure}[h]
    \centering
    \includegraphics[width=0.4\linewidth]{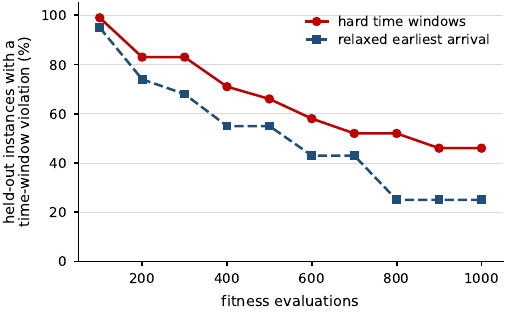}
    \caption{Fraction of held-out routing instances on which the best heuristic violates a time-window constraint, against fitness evaluations spent.
    The dashed curve relaxes the earliest-arrival bound so that vehicles may serve customers early.
    Neither curve is decreasing at the end of the budget.}
    \label{fig:vrptw}
\end{figure}

\paragraph{Violations Against Search Budget.}
Figure \ref{fig:vrptw} tracks time-window violations along the search trajectory of Section \ref{subsec:ladder}.

\begin{figure}[h]
    \centering
    \includegraphics[width=0.50\linewidth]{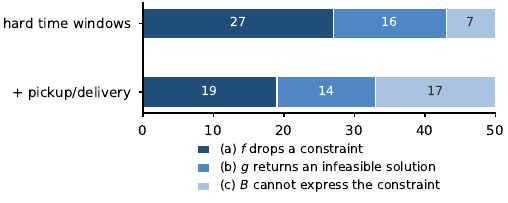}
    \caption{Failure mechanism for $50$ sampled infeasible instance-transformation pairs at each of the two tight rungs of the constraint ladder, classified by hand.}
    \label{fig:taxonomy}
\end{figure}

\paragraph{How Transformations Fail.}
Figure \ref{fig:taxonomy} gives the counts behind the taxonomy of Section \ref{subsec:taxonomy}.

\clearpage
\paragraph{Standard Versus Obfuscated Descriptions.}
Table \ref{tab:retrieval} reports the experiment of Section \ref{subsec:retrieval} in full.
On the two knapsack-family problems the degradation is nil or slightly positive, within run-to-run noise, and on TSP it is small ($1.4\%$), whereas on CVRP it is $5.9\%$ and on the last-mile task feasibility falls from $61.3\%$ to $38.5\%$.
Selecting items under a budget is recognizable however it is phrased, and a partitioning of customers into capacity-feasible routes is not.

\begin{table}[h]
    \centering
    \footnotesize
    \caption{Standard versus obfuscated problem descriptions.
    The obfuscated descriptions are mathematically equivalent, use no standard terminology or problem names, and were written before any run.
    Mean over five runs.
    The last-mile column reports feasibility rather than objective, since that is what fails there.}
    \begin{tabular}{l|cc|r}
        \thickhline
        Problem (metric) & Standard & Obfuscated & $\Delta$ \\\hline
        TSP $n$=50 (Obj. $\downarrow$) & 5.819 & 5.902 & $+1.4\%$ \\
        CVRP $n$=50 (Obj. $\downarrow$) & 9.826 & 10.402 & $+5.9\%$ \\
        KP $n$=50 (Obj. $\uparrow$) & 20.006 & 19.947 & $-0.3\%$ \\
        MKP $n$=100 (Obj. $\uparrow$) & 23.164 & 22.913 & $-1.1\%$ \\
        BPP $n$=500 (Obj. $\downarrow$) & 203.344 & 205.107 & $+0.9\%$ \\\hline
        Last-mile (Feas. $\uparrow$) & 61.3\% & 38.5\% & $-22.8$\,pt \\
        Valid at init., mean (\%\,$\uparrow$) & 78.3\% & 51.2\% & $-27.1$\,pt \\
        \thickhline
    \end{tabular}
    \label{tab:retrieval}
\end{table}



\subsection{Examples of Designed Transformations and Heuristics from ReFrame}\label{subapp:demo}

\paragraph{A Worked Example.}
We now trace one run of the framework from its handcrafted input to the heuristic it returns.
For the capacitated vehicle routing problem, the handcrafted input is a five-sentence description (customers with demands, a depot, vehicles of capacity $C$, and the objective of minimizing total distance) together with a predicate checking that every customer appears exactly once and that no route exceeds $C$.
From this, one run proposed among its ten candidates a target problem described as ``partitioning $n$ weighted points into groups whose total weight does not exceed a bound, then ordering the points within each group to minimize travel'', which is capacitated clustering followed by per-cluster sequencing.
Its instance map $f$ emitted the coordinate matrix, the demand vector, and $C$ unchanged, and its solution map $g$ concatenated the per-cluster orderings into routes separated by depot visits.
Its code template asked for a function taking the distance matrix, demands, and capacity and returning a list of groups with an ordering for each.
The evolved heuristic filled that signature with a savings-style seeding of clusters followed by a cheapest-insertion ordering, and later acquired a $2$-opt pass over each cluster's ordering through a parent drawn from a different target problem's population.
Two other transformations survived selection in the same run: a giant-tour formulation split into capacity-feasible segments, and an assignment formulation that balanced load across a fixed vehicle count before sequencing.
None of the three is novel as an idea.
The point is that none of them was written by us, and that each arrived with the signature and evaluator harness that the search needs.
The transformations selected across all problems and runs are reported in Table~\ref{tab:chosen}.

\clearpage
\paragraph{MKP\@.}
Figures \ref{fig:learned-transform-mkp} and \ref{fig:learned-heuristic-mkp} respectively show an example of the designed transformation and the corresponding heuristic for MKP.

\begin{figure}[h]
    \footnotesize
    \centering
    \begin{tcolorbox}[width=\textwidth, title={Problem Description}, fonttitle=\bfseries\footnotesize]
        Problem B1 involves selecting a subset of N items such that the total value is maximized while ensuring that the total weights in each of the M dimensions do not exceed specific limits, using a greedy heuristic approach based on value-to-weight ratios.
    \end{tcolorbox}
    \begin{lstlisting}[captionpos=b]
import numpy as np
from typing import Tuple, List

def convert_input_A_to_B(values: np.ndarray, weights: np.ndarray, constraints: np.ndarray) -> Tuple[np.ndarray, np.ndarray, np.ndarray]:
    ''' Convert input of Problem A into input of Problem B
    Args:
    values (np.ndarray): A 1D float array of length N storing the item values.
    weights (np.ndarray): A (M x N) float matrix storing the multi-dimensional weights, where each row is associated with a constraint.
    constraints (np.ndarray): A 1D float array of length M storing weight constraints.

    Returns:
    input_B: A tuple storing the corresponding input of Problem B.
    '''
    # Calculate value-to-weight ratios for each item
    ratios = values / np.sqrt(np.sum(np.square(weights), axis=0))  # Changed to root of sum of squares for better ratio
    input_B = (values, weights, constraints, ratios)

    return input_B


def convert_solution_B_to_A(solution_B: List[int]) -> List[int]:
    ''' Convert solution of Problem B into solution of Problem A
    Args:
    solution_B: The output of Problem B, which contains indices of selected items.

    Returns:
    items: A list storing the indices of selected items subject to the weight constraints.
    '''
    items = list(solution_B)
    return items
    \end{lstlisting}
    \begin{lstlisting}[captionpos=b]
def convert_input_A_to_B(values, weights, constraints):
    ''' Convert input of Problem A into input of Problem B
    Args:
    values (np.ndarray): A 1D float array of length N storing the item values.
    weights (np.ndarray): A (M x N) float matrix storing the multi-dimensional weights, where each row is associated with a constraint.
    constraints (np.ndarray): A 1D float array of length M storing weight constraints.

    Returns:
    input_B: A tuple storing the corresponding input of Problem B.
    '''
    # Calculate value-to-weight ratios for each item
    ratios = values / np.sum(weights, axis=0)
    input_B = (values, weights, constraints, ratios)

    return input_B


def convert_solution_B_to_A(solution_B):
    ''' Convert solution of Problem B into solution of Problem A
    Args:
    solution_B: The output of Problem B.

    Returns:
    items: A list storing the indices of selected items subject to the weight constraints.
    '''
    selected_items = solution_B  # Assuming solution_B contains the indices of selected items
    items = list(selected_items)

    return items
    \end{lstlisting}
    \caption{Designed transformation for MKP using ReFrame. (Top) Problem description of $B$, (center) refined and (bottom) original implementation of $(f,g)$ for transforming MKP to $B$.}
    \label{fig:learned-transform-mkp}
\end{figure}

\begin{figure}
    \ContinuedFloat
    \captionsetup{list=off,format=cont}
    \footnotesize
    \centering
    \begin{lstlisting}[captionpos=b]
from typing import Tuple, List

def solve_B(input_B: Tuple[np.ndarray, np.ndarray, np.ndarray, np.ndarray]) -> List[int]:
    '''
    Args:
    input_B (Tuple[np.ndarray, np.ndarray, np.ndarray, np.ndarray]): A tuple containing:
        - values (np.ndarray): A 1D float array of length N storing the item values.
        - weights (np.ndarray): A (M x N) float matrix storing the multi-dimensional weights, where each row is associated with a constraint.
        - constraints (np.ndarray): A 1D float array of length M storing weight constraints.
        - ratios (np.ndarray): A 1D float array of length N storing the value-to-weight ratios for each item.

    Returns:
    List[int]: A list storing the indices of selected items subject to the weight constraints.
    '''

    return solution_B
    \end{lstlisting}
    \caption{Code template for solving $B$.}
\end{figure}

\begin{figure}[h]
    \footnotesize
    \centering
    \begin{tcolorbox}[width=\textwidth, title={Problem Description}, fonttitle=\bfseries\footnotesize]
        Problem B1 involves selecting a subset of N items such that the total value is maximized while ensuring that the total weights in each of the M dimensions do not exceed specific limits, using a greedy heuristic approach based on value-to-weight ratios.
    \end{tcolorbox}
    \begin{tcolorbox}[width=\textwidth, title={Heuristic Description}, fonttitle=\bfseries\footnotesize]
        A new algorithm that selects items iteratively, calculating the best score considering both value and the remaining capacity left in multi-dimensional space, while simultaneously updating the constraints as items are selected.
    \end{tcolorbox}
    \begin{lstlisting}[captionpos=b]
from typing import Tuple, List
import numpy as np

def solve_B(input_B: Tuple[np.ndarray, np.ndarray, np.ndarray, np.ndarray]) -> List[int]:
    values, weights, constraints, ratios = input_B
    M, N = weights.shape
    selected_items = []
    total_weights = np.zeros(M)

    # Calculate the remaining capacity to define the score more effectively
    remaining_capacity = constraints.copy()

    while True:
        best_score = -np.inf
        best_item = -1
        for idx in range(N):
            if idx in selected_items:
                continue
            item_weight = weights[:, idx]

            if all(total_weights + item_weight <= constraints):
                # Calculate new score based on value and remaining capacity
                score = values[idx] / (np.sum(item_weight / remaining_capacity) + 1e-9)  # Avoid division by zero
                if score > best_score:
                    best_score = score
                    best_item = idx
        
        if best_item == -1:  # No feasible item can be added
            break

        selected_items.append(best_item)
        total_weights += weights[:, best_item]
        remaining_capacity -= weights[:, best_item]

    return selected_items
    \end{lstlisting}
    \caption{Designed heuristic using the transformation for MKP in Figure \ref{fig:learned-transform-mkp}.}
    \label{fig:learned-heuristic-mkp}
\end{figure}

\clearpage
\paragraph{TSP\@.}

Figure \ref{fig:learned-heuristic-tsp} shows the learned heuristic whose performance was reported in Table \ref{tab:tsp-sbs}.

\begin{figure}[h]
    \footnotesize
    \centering
    \begin{tcolorbox}[width=\textwidth, title={Problem Description}, fonttitle=\bfseries\footnotesize]
        Problem B7 involves applying a nearest neighbor heuristic starting from each node and selecting the route with the least total distance as a near-optimal solution.
    \end{tcolorbox}
    \begin{tcolorbox}[width=\textwidth, title={Heuristic Description}, fonttitle=\bfseries\footnotesize]
        Repeatedly select a random starting point and build a route while ensuring that each node is added without forming a local loop, then refine the tour using a hill-climbing strategy to find better routes.
    \end{tcolorbox}
    \begin{lstlisting}[captionpos=b]
from typing import Tuple
import numpy as np

def solve_B(input_B: Tuple[np.ndarray]) -> np.ndarray:
    '''
    Args:
    input_B (Tuple[np.ndarray]): A tuple containing a single np.ndarray, which is a NxN matrix where the entry
                                 at i-th row and j-th column (or vice versa) stores the Euclidean distance between
                                 nodes i and j.

    Returns:
    np.ndarray: A 1D array of length N containing the unique node IDs to visit in order, representing the solution
                of Problem B converted to the corresponding solution of Problem A.
    '''
    distances = input_B[0]
    num_nodes = distances.shape[0]
    best_route = []
    best_length = float('inf')

    # {Repeatedly select a random starting point and build a route while ensuring that each node is added without forming a local loop,
    # then refine the tour using a hill-climbing strategy to find better routes.}
    for _ in range(10):  # Number of iterations to random start points
        start_node = np.random.randint(num_nodes)
        visited = [start_node]
        unvisited = set(range(num_nodes))
        unvisited.remove(start_node)

        while unvisited:
            current_node = visited[-1]
            next_node = min(unvisited, key=lambda x: distances[current_node][x])
            visited.append(next_node)
            unvisited.remove(next_node)

        # Hill-climbing optimization
        improvement = True
        while improvement:
            improvement = False
            for i in range(len(visited)):
                for j in range(i + 2, len(visited)):
                    if j == len(visited) - 1 and i == 0:  # Skip the final edge to start
                        continue
                    current_cost = (distances[visited[i]][visited[(i + 1) % len(visited)]] +
                                     distances[visited[j]][visited[(j + 1) % len(visited)]])
                    new_cost = (distances[visited[i]][visited[j]] +
                                 distances[visited[(i + 1) % len(visited)]][visited[(j + 1) % len(visited)]])

                    if new_cost < current_cost:
                        visited[i + 1:j + 1] = reversed(visited[i + 1:j + 1])
                        improvement = True
                        break
                if improvement:
                    break

        current_length = sum(distances[visited[k]][visited[(k + 1) % len(visited)]] for k in range(len(visited)))

        if current_length < best_length:
            best_length = current_length
            best_route = visited

    return np.array(best_route)
    \end{lstlisting}
    \caption{Designed heuristic for TSP using ReFrame.}
    \label{fig:learned-heuristic-tsp}
\end{figure}

\clearpage

\subsection{Setup for the Constraint Ladder and the Last-Mile Task}\label{subapp:limit}

This appendix records the instance-construction details behind Sections \ref{subsec:lastmile} and \ref{subsec:ladder}.

\paragraph{The Constraint Ladder.}
All four rungs share the CVRP instance geometry of Section \ref{subsec:setup}: $n=50$ customers with coordinates drawn uniformly in the unit square, integer demands drawn uniformly from $\{1,\dots,9\}$, and vehicle capacity $C=50$.
Rung~0 is that problem unchanged.
Rung~1 attaches a time window $[t_i^{\mathrm{start}},t_i^{\mathrm{end}}]$ to each customer and charges a linear penalty per unit of lateness, added to the objective, so that every solution remains feasible by construction.
Rung~2 makes the same windows hard, so that a route arriving outside $[t_i^{\mathrm{start}},t_i^{\mathrm{end}}]$ is infeasible.
Windows are generated with the procedure of \citet{liu2024llm4ad}, which places each window around a feasible arrival time on a reference solution so that a feasible solution is guaranteed to exist.
Rung~3 additionally pairs customers into pickup-and-delivery couples, requiring the pickup to precede its delivery on the same route.
Training uses $64$ instances per rung and testing another $64$.
The framework settings, the search budget, and the designer LLM are identical across rungs, so the only variable is the shape of the feasible set.

\paragraph{The Last-Mile Instances.}
From the 2021 Amazon Last Mile Routing Research Challenge data set \citep{merchan2024amazon} we take the Austin and Boston route subsets and sample $120$ routes uniformly, discarding routes with fewer than $60$ or more than $250$ stops.
Each route yields one instance consisting of its stop set, the travel-time matrix distributed with the data, per-stop service times, the time windows attached to packages at that stop (stops whose packages carry no window are treated as unconstrained), and a capacity taken as the total package volume of the original route.
Sixty-four instances are used during the search and the remaining $56$ are held out.
The Gehring--Homberger instances \citep{gehring1999parallel} are used as distributed, at $200$ customers, across the C2, R2 and RC2 families.

\paragraph{Trivial Transformations.}
We record one failure mode separately because it is easy to mistake for a bug.
When the designer LLM has no useful domain knowledge for $A$, it sometimes proposes a transformation in which $f$ returns its input unchanged and $g$ returns the heuristic's output unchanged, so that the evolutionary search is effectively run on $A$ directly with no specification at all.
These candidates are not invalid, since they satisfy Definition \ref{def:transformation} trivially, and they are not always discarded by selection, because on loosely constrained problems they score adequately.
We observed them in $6$ of the $350$ candidates generated across the seven problems, all on the last-mile task and on the obfuscated descriptions of Section \ref{subsec:retrieval}.
The practical response is the one recommended in Section \ref{sec:limit}, namely to use a stronger model for initialization, where it is cheap.

\end{document}